\documentclass[lettersize,journal]{IEEEtran}
\usepackage{amsmath,amsfonts}
\usepackage{algorithmic}
\usepackage{array}
\usepackage[caption=false,font=normalsize,labelfont=sf,textfont=sf]{subfig}
\usepackage{textcomp}
\usepackage{stfloats}
\usepackage{url}
\usepackage{verbatim}
\usepackage{graphicx}
\usepackage{footnote}
\usepackage{longtable}
\usepackage{adjustbox}
\def\BibTeX{{\rm B\kern-.05em{\sc i\kern-.025em b}\kern-.08em
    T\kern-.1667em\lower.7ex\hbox{E}\kern-.125emX}}
\usepackage{balance}
\usepackage{cite}
\usepackage{tabularx}
\usepackage{multirow}
\usepackage{makecell}
\usepackage{pifont}
\usepackage{titlesec}
\titlespacing{\section}{0pt}{2.5ex plus .2ex minus .2ex}{1.0ex plus .2ex}
\begin{document}
\newcommand{\e}[1]{\textbf{#1}}
\title{FD-SLAM: Fast Dense Radar-Inertial SLAM with Frequency-Domain Loop Closure and Pose Graph Optimization}
\author{Nader J. Abu-Alrub, Nathir A. Rawashdeh}

\maketitle

\begin{abstract}

Radar SLAM is attractive for autonomous ground vehicles operating in visually degraded environments, however, scanning radars are noisy, have low scanning rates, and their measurements are challenging to match reliably over long trajectories. This paper presents FD-SLAM, a fast dense radar-inertial SLAM system that extends dense radar-inertial odometry with frequency-domain loop closure and pose graph optimization. The proposed method preserves an image-like structure of scanning radar measurements by using a compact frequency-domain polar descriptor for loop-candidate retrieval and a multi-stage verification pipeline based on temporal filtering, phase-correlation screening, scan-alignment similarity, and geometric consistency checks. Verified loop closures are added as non-sequential constraints in an SE(2) pose graph together with radar-inertial odometry factors. FD-SLAM is evaluated on a publicly available dataset using standard KITTI evaluation metrics. The results show that FD-SLAM improves FD-RIO baseline, achieves competitive performance against current state-of-the-art radar SLAM methods, and provides favorable rotational accuracy across multiple evaluated driving trajectories. Runtime analysis further indicates that the radar-inertial front-end operates above the radar sampling rate on a CPU-only setup, while loop closure detection and graph optimization remain suitable for parallel background execution.

\end{abstract}


\section{Introduction}\label{sec:introduction}

\begin{figure}[!t]
    \centering
    \subfloat[]{\includegraphics[width=3.0in]{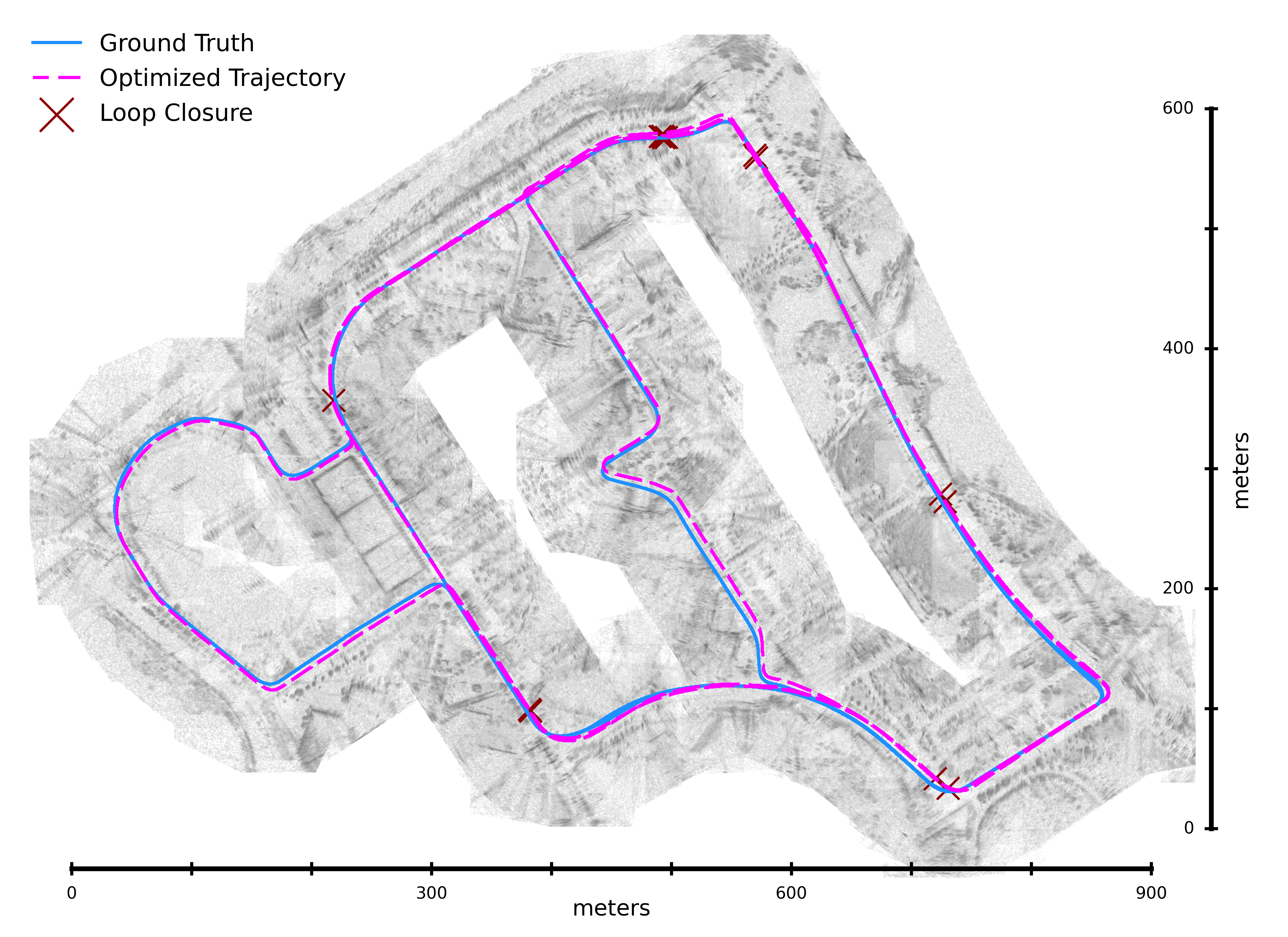}\label{fig:trajectory_and_map}}\hfil
    \subfloat[]{\includegraphics[width=3.0in]{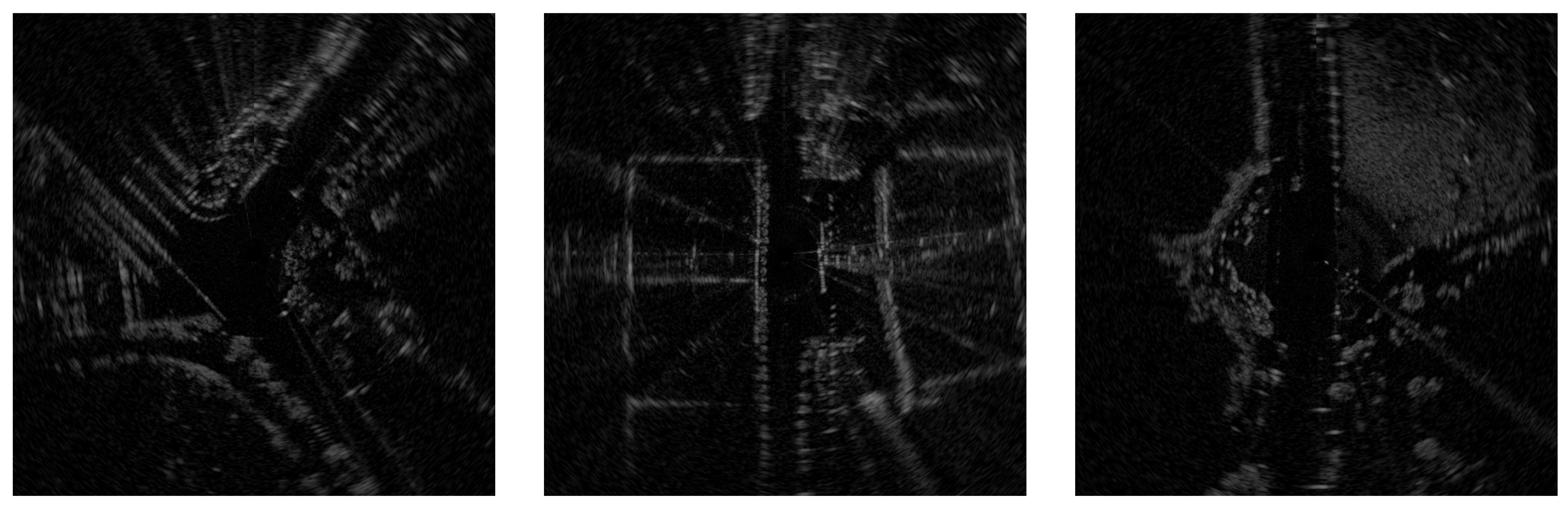}\label{fig:sample_scans}}
    \caption{Optimized FD-SLAM trajectory overlaid on an accumulated radar map. (a) The optimized trajectory is shown over the radar map generated for sequence \textit{KAIST03}. (b) Sample radar scans from the same sequence, illustrating the image-like structure of the scans and the presence of noise and artifacts.}
    
\end{figure}

\IEEEPARstart{S}{imultaneous} localization and mapping (SLAM) is a fundamental capability for autonomous ground vehicles operating in unknown or GPS-denied environments. A SLAM system must estimate the vehicle trajectory while maintaining a consistent representation of the surrounding environment. This capability is essential for long-term navigation, map construction, and autonomous operation in urban, suburban, and off-road environments. While camera- and LiDAR-based SLAM systems have achieved strong performance in many scenarios, their reliability can degrade under poor illumination, adverse weather, airborne particles, and other visibility-limited conditions \cite{zhang2023weather1, mohammed2020weather2, bijelic2018weather3, RadarVSLidar4Loc}. Radar sensing is therefore increasingly attractive for autonomous navigation because millimeter-wave radar can operate in darkness, fog, dust, rain, and snow, where other sensors may suffer from reduced reliability.

Among radar sensors, mechanically scanning frequency-modulated continuous-wave (FMCW) radar is particularly useful for ground-vehicle localization because it provides dense, long-range, 360-degree bird's-eye-view measurements of the environment. Unlike sparse automotive radar point clouds, measurements from scanning radar can be formatted as image-like intensity scans that preserve global spatial structure around the vehicle. However, these measurements also introduce significant challenges. Radar scans are noisy, contain multipath reflections and speckle artifacts, and are affected by motion distortion due to the finite scan acquisition time. In addition, scanning radar typically operates at relatively lower sampling rates compared with other sensors used on autonomous vehicles. As a result, radar-only odometry can accumulate drift over long trajectories, while scan matching can be difficult in feature-poor or perceptually aliased environments.

Recent radar SLAM systems have shown that radars can support large-scale localization and mapping in challenging conditions. However, many existing systems convert radar measurements into sparse point sets, landmarks, or feature representations before performing odometry, place recognition, or loop closure verification. Such representations can be effective, but they may discard useful information contained in the dense radar image. Other systems adapt descriptors or verification mechanisms originally developed for LiDAR or camera pipelines. This motivates the question: can the dense image-like structure of scanning radar be preserved and used throughout the SLAM pipeline, from local motion estimation to loop-candidate retrieval, loop verification, and global pose graph correction?

This paper addresses that question by introducing FD-SLAM, a fast dense radar-inertial SLAM system for scanning radar and IMU sensing. FD-SLAM extends our previous Fast Dense Radar-Inertial Odometry (FD-RIO)\cite{NaderFDRio} front-end into a complete graph-based SLAM pipeline. The radar-inertial front-end provides local motion estimates by combining dense correlation-based radar odometry with high-rate IMU measurements. These local estimates are inserted into a pose graph as sequential odometry constraints. In parallel, each radar scan is converted into a compact descriptor derived from the Fourier magnitude and polar frequency representation of the scan. Descriptor similarity is used to retrieve candidate revisits, while temporal filtering, phase-correlation screening, scan-alignment similarity, relative transform estimation, and geometric consistency checks are used to verify loop closures before adding them to the graph. An example FD-SLAM result is shown in Fig. \ref{fig:trajectory_and_map}, where the optimized trajectory is overlaid on an accumulated radar map generated from the scanning radar measurements. While Fig. \ref{fig:sample_scans} illustrates sample scans taken from the same trip.

The proposed system is designed around a dense radar-native representation. Rather than reducing each radar scan to sparse landmarks or point detections, FD-SLAM preserves the scan as an image-like signal and exploits its frequency-domain structure. The resulting pose graph combines sequential radar-inertial odometry factors with verified non-sequential loop closure factors, allowing global trajectory drift to be reduced while preserving the local consistency of the odometry front-end. The main contributions of this paper are as follows:

\begin{itemize}
    \item A complete dense radar-inertial SLAM pipeline that extends correlation-based radar-inertial odometry with descriptor-based loop closure detection and pose graph optimization.
    \item A compact frequency-domain polar radar descriptor and a deterministic multi-stage loop closure verification pipeline that preserves dense scanning radar structure and enables efficient candidates retrieval over long trajectories.
    \item Evaluation on publicly available dataset showing that FD-SLAM is on par with the current state-of-the-art methods in terms of translational and rotational accuracy, in addition to runtime capabilities.
\end{itemize}

The remainder of this paper is organized as follows. Section \ref{sec:related_work} reviews related work in radar odometry and radar SLAM. Section \ref{sec:proposed_method} presents the proposed FD-SLAM system, including the radar-inertial odometry front-end, scan matching and loop closure module, and pose graph optimization back-end. Section \ref{sec:evaluation} evaluates the proposed system qualitatively and quantitatively on a publicly available dataset and analyzes descriptor behavior and runtime performance. Section \ref{sec:conclusion} concludes the paper and discusses future directions.

\section{Related Work}\label{sec:related_work}
\noindent
Radar-based odometry and SLAM have received increasing attention for autonomous ground vehicles because radar can provide robust perception under adverse illumination and weather conditions where cameras and LiDAR may degrade. Radar odometry estimates the motion of a platform relative to an initial frame, while SLAM additionally estimates a globally consistent trajectory and map by detecting revisited locations and correcting accumulated drift. Radar odometry methods are generally categorized as sparse, dense, or hybrid approaches. Sparse methods rely on extracted detections, landmarks, or point sets; dense methods operate directly on radar intensity images or scan representations; and hybrid methods combine aspects of both. This distinction is particularly important for spinning FMCW radar, which provides dense 360° bird's-eye-view scans but also introduces challenges such as speckle noise, multipath reflections, motion distortion, and lower scan rates \cite{NaderSurvey}.

Early radar SLAM work demonstrated that radar scans could be used directly for localization and mapping without relying on conventional visual or LiDAR-style landmarks. Checchin \textit{et al.} proposed a radar scan-matching SLAM method based on the Fourier-Mellin Transform, using global frequency-domain image registration to estimate the relative rotation and translation between radar scans \cite{checchin2010radar}. Their work treats radar scans as dense images and avoids explicit landmark extraction and data association. However, the work used an earlier custom FMCW radar platform and reported real-time data acquisition with offline SLAM processing. Related radar-only localization and mapping methods, such as ROLAM and subsequent evaluations of radar mapping approaches, further explored radar-specific scan distortion and dense registration ideas, but were developed before modern large-scale radar datasets and recent graph-based radar SLAM pipelines \cite{vivet2013mobile}. More recent dense radar odometry methods have revisited these ideas using modern scanning radar. PhaRaO, for example, uses phase correlation and Fourier-Mellin registration for direct radar odometry, showing that relative motion can be estimated effectively by exploiting the Fourier-domain structure of radar scans rather than extracting sparse features \cite{park2020pharao}. These works motivate the radar-native design philosophy followed in this paper: radar scans are treated as dense image-like measurements whose frequency-domain structure can be directly exploited for registration and loop-closure verification.

A number of complete radar SLAM systems have been proposed using 2D spinning radar. RadarSLAM by Hong \textit{et al.} is one of the earliest large-scale full radar SLAM systems, combining radar motion estimation, loop closure detection, and pose graph optimization \cite{hong2022radarslam}. Their front-end detects and tracks radar keypoints using image-processing and feature-tracking techniques, while loop closure is performed by converting radar polar scans into point clouds and using an adapted M2DP descriptor before inserting loop constraints into a graph. RadarSLAM demonstrated the feasibility of large-scale radar SLAM under adverse weather, but its loop-closure pipeline relies on extracted radar features and a descriptor originally developed for point-cloud place recognition. This differs from dense radar-native approaches, where the radar image representation is preserved throughout descriptor extraction, candidate verification, and relative transform estimation.

TBV Radar SLAM by Adolfsson \textit{et al.} combines CFEAR radar odometry \cite{adolfsson2023cfear}, Scan Context-based place recognition, loop-candidate registration, introspective verification, and pose graph optimization \cite{adolfsson2023tbv}. Its main contribution is a “trust but verify” loop-closure strategy in which multiple candidates are retrieved and then verified using place similarity, odometry consistency, and learned alignment-quality measures before being accepted as graph constraints. TBV operates primarily on sparse radar representations: CFEAR filters the radar scan into peak detections and oriented surface points, and the loop descriptor is constructed from aggregated, motion-compensated radar peaks rather than directly from the full dense radar image. The method achieves strong translational accuracy, but its design differs from the present work in several important ways. In contrast to TBV, the proposed FD-SLAM system preserves the dense radar scan structure and performs loop-candidate retrieval and verification using FFT/polar descriptors, phase-correlation screening, scan-alignment similarity, and geometric consistency checks. The proposed system also uses a radar-inertial odometry front-end and avoids a learned loop-verification module.

More recently, Dr-PoGO introduced a direct radar pose-graph optimization framework for 2D spinning FMCW radar \cite{gentil2026dr}. Dr-PoGO uses direct radar odometry, RaPlace-based loop-candidate detection, coarse-to-fine loop-closure registration, and SE(2) pose graph optimization. It is especially relevant because it also performs graph-based SLAM with dense scanning radar. However, Dr-PoGO differs in its loop-closure design. Loop candidates are detected using RaPlace descriptors computed from local maps, and relative loop transformations are initialized using SIFT feature matching and RANSAC before being refined by direct cross-correlation. Therefore, although Dr-PoGO uses direct registration for odometry and final loop refinement, it still relies on visual features for coarse loop-closure initialization. In contrast, FD-SLAM avoids visual feature descriptors and performs candidate retrieval and verification through radar-native frequency-domain and correlation-based stages. In addition, Dr-PoGO performs loop closure between local maps generated by direct radar odometry, whereas the proposed method uses scan-level dense radar descriptors and verification before adding loop factors to the graph.

Radar place recognition is closely related to loop closure, but it does not by itself constitute a full SLAM solution. The \textit{MulRan} dataset was introduced to support range-based urban place recognition using LiDAR and scanning radar, and it provides repeated trajectories with a Navtech CIR204-H radar, making it highly relevant for evaluating radar loop-closure methods \cite{mulran_dataset}. The same work introduced Radar Scan Context, an adaptation of Scan Context to radar polar images, showing that radar can provide effective place recognition in urban environments. Other radar place-recognition methods have explored learned rotationally invariant radar embeddings, sequence-based matching, contrastive learning, Radon-transform descriptors, and heterogeneous radar-to-LiDAR retrieval \cite{gadd2020look,yin2021radar,jang2023raplace}. These works demonstrate that radar descriptors can support candidate retrieval under challenging conditions.

Radar SLAM has also been investigated using sparse automotive and single-chip radar sensors. Holder \textit{et al.} proposed a real-time pose graph SLAM system using a single front-facing automotive radar, building submaps from sparse radar detections, performing scan matching, detecting loop closures, and optimizing a pose graph \cite{holder2019real}. More recent automotive radar systems exploit radar-specific measurements such as Doppler velocity and radar cross section. RaI-SLAM, for example, combines automotive radar and IMU measurements using local and global pose graphs, and incorporates loop closure detection using radar-specific cues \cite{herraez2025rai}. Radarize instead targets indoor SLAM with low-cost, off-the-shelf single-chip mmWave radar, using Doppler-based translation estimation, learned rotation estimation from radar heatmaps, artifact rejection, and a Cartographer back-end \cite{sie2024radarize}. Other array or snapshot radar SLAM systems use angular super-resolution or submap-based ICP to compensate for the limited angular resolution of low-cost radar hardware. These methods show the breadth of radar SLAM research beyond spinning radar, but they address a different sensing regime: sparse, limited-field-of-view radar detections or heatmaps rather than dense 360° scanning-radar imagery.

A separate line of recent work has explored 4D imaging radar SLAM. Unlike 2D spinning radar, 4D radar provides sparse 3D point clouds with Doppler velocity information, usually over a limited field of view. 4DRadarSLAM uses 4D radar point-cloud registration, intensity Scan Context loop detection, and pose graph optimization for large-scale environments \cite{zhang20234dradarslam}. Other methods introduce Doppler-derived ego-velocity preintegration factors, radar-inertial-velocity graph formulations, ground factors, and introspective loop-closure verification for 4D radar \cite{li20234d,wang2024riv,hilger2025introspective}. These systems demonstrate the rapid growth of graph-based radar SLAM with modern imaging radar sensors. However, they solve a different problem from the one addressed in this paper: sparse 3D/4D radar point-cloud SLAM rather than dense 2D scanning-radar SLAM.

To summarize, the proposed FD-SLAM system contributes a complementary design point: a dense radar-inertial SLAM pipeline that preserves the native scanning-radar image representation, uses a compact FFT/polar descriptor for loop-candidate retrieval, verifies candidates through deterministic tests, and incorporates accepted loop closures into a pose graph for global trajectory correction. This work demonstrates that dense frequency-domain radar representations can support a complete SLAM pipeline from local odometry to loop closure and graph optimization.

\section{Proposed Method}\label{sec:proposed_method}

\subsection{Overview}\label{subsec:overview}

\begin{figure*}[!t]
    \centering
    \includegraphics[width=7.0in]{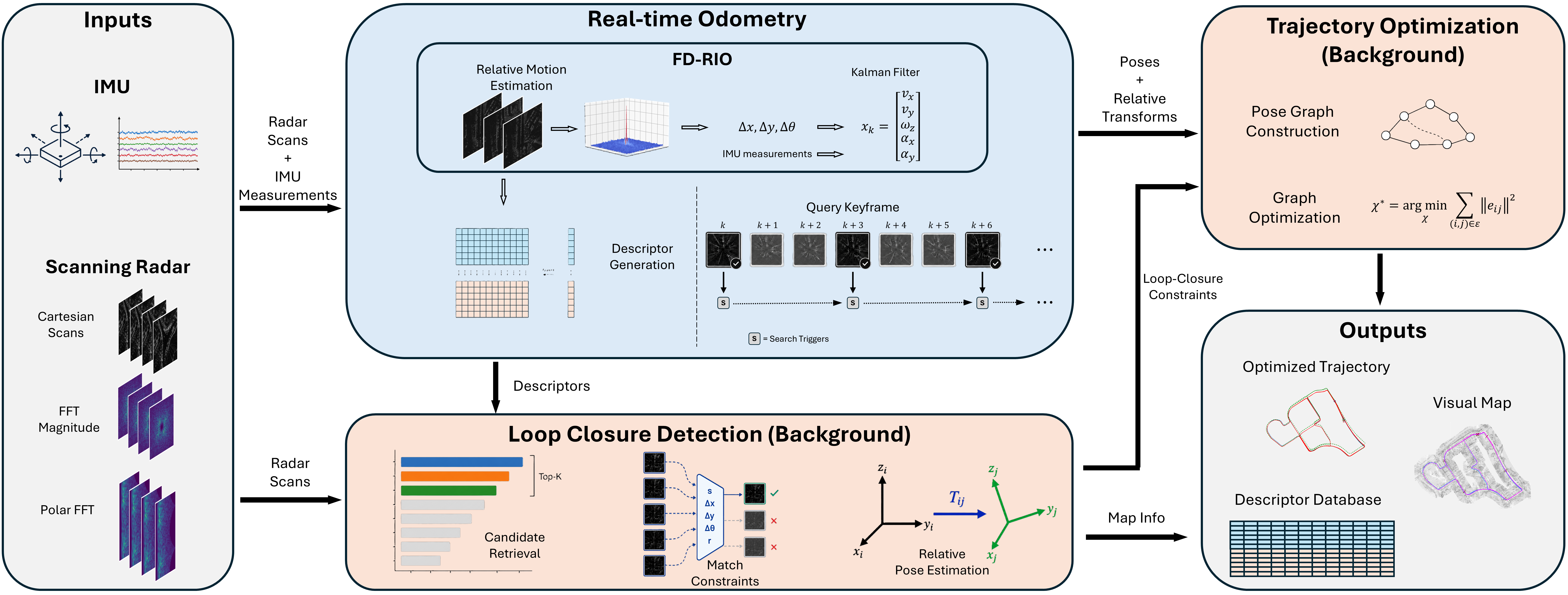}
    \caption{Overview of FD-SLAM method. Figure shows the main components of the system and their interactions. The radar-inertial odometry front-end generates local motion estimates between consecutive poses, which are used to construct a pose graph. Loop closure detection identifies previously visited locations and adds non-sequential constraints to the graph. The back-end optimizes the full pose graph to produce a globally consistent trajectory estimate.}
    \label{fig:system_overview}
\end{figure*}

The proposed FD-SLAM system extends our radar-inertial odometry front-end into a complete graph-based SLAM pipeline. The system is designed around three main components as illustrated in Fig. \ref{fig:system_overview}: radar-inertial odometry, radar scan matching for loop closure detection, and pose graph optimization. The radar-inertial odometry front-end provides locally consistent motion estimates between consecutive radar scans, while the loop closure module identifies revisited locations and estimates non-sequential relative constraints. These constraints are then incorporated into a pose graph, where optimization is used to reduce accumulated drift and improve the global consistency of the trajectory. The loop closure detection module and trajectory optimization module are run in the background using two separate processes; this design allows the radar-inertial odometry front-end to operate in real time without being affected by the computational load of loop closure and graph optimization.

The system receives radar scans and IMU measurements as inputs. The IMU measurements are processed at high rate by the radar-inertial odometry front-end, while radar scans are processed using a dense correlation-based registration method. The output of this front-end is a sequence of relative motion estimates, which are integrated into poses and added to the pose graph as odometry factors between consecutive nodes. Each node in the graph represents a planar vehicle pose, and each sequential edge represents the relative motion estimated by the odometry front-end.

In parallel with odometry estimation, each radar scan is also converted into a compact descriptor for loop closure search. The descriptor database is incrementally updated as new scans are processed. For a given query scan, descriptor cosine similarity is used to retrieve a small subset of candidate revisits from previous scans. Temporally close candidates are removed to avoid selecting neighboring scans along the trajectory. The remaining candidates are then passed through a sequence of verification stages, including phase-correlation screening, normalized scan-alignment similarity, and geometric consistency checks on the estimated relative transform. This multi-stage process is used to reduce the likelihood of false loop closure constraints before they are inserted into the pose graph.

When a candidate loop closure passes all verification stages, the estimated relative transform between the current scan and the matched previous scan is added to the graph as a non-sequential loop closure factor. The pose graph therefore contains two types of relative constraints: odometry factors from consecutive radar-inertial updates and loop closure factors from verified revisits. A prior factor is added to the first pose to anchor the graph. The optimized trajectory is obtained by solving the resulting nonlinear least-squares problem over all pose nodes.

The proposed architecture separates local motion estimation from global trajectory correction. The radar-inertial odometry front-end is responsible for real-time local pose propagation, while the loop closure and graph optimization modules correct long-term drift when revisited locations are detected. This modular design allows the system to preserve the computational efficiency of the dense radar-inertial odometry front-end while improving global consistency through descriptor-based loop closure detection and pose graph optimization.

\subsection{Radar-Inertial Odometry}\label{subsec:radar_inertial_odometry}

FD-SLAM uses our previously developed Fast Dense Radar-Inertial Odometry (FD-RIO) method \cite{NaderFDRio} as its front-end. In this work, FD-RIO serves as the local motion estimation module that provides sequential constraints for the SLAM back-end. Only a brief summary is given here, while full implementation and evaluation details can be found in \cite{NaderFDRio}.

FD-RIO combines dense scanning radar odometry with inertial measurements from an IMU. The radar component estimates the relative motion between consecutive radar scans using a correlation-based registration pipeline. In particular, the relative rotation is estimated in the Fourier domain using the magnitude spectra of the radar scans in polar form, after which the rotationally aligned scans are used to estimate translation in the spatial domain. This dense registration strategy avoids explicit feature extraction and instead operates directly on the image-like structure of the radar scans.

The resulting radar motion estimates are fused with IMU measurements in a Kalman filter. The filter maintains planar motion states including linear velocity, yaw rate, and linear acceleration, and is updated asynchronously as radar and IMU measurements arrive. The IMU provides high-rate motion information, while the radar measurements provide exteroceptive motion estimates that help constrain drift and correct the filter state. In this way, FD-RIO exploits the complementary properties of both sensing modalities: radar provides robustness in challenging visibility conditions, while the IMU provides high-rate motion updates between radar scans.

The fused velocity estimates are integrated over time to recover the local pose trajectory. For the SLAM system, the relative transformation between consecutive poses is used as the odometry measurement between neighboring graph nodes. These sequential odometry constraints form the backbone of the pose graph, while the SLAM back-end later introduces loop closure constraints to reduce accumulated drift and improve global consistency.

\subsection{Scan Matching and Loop Closure}\label{subsec:scan_matching_loop_closure}

Loop closure detection is used to identify previously visited locations and to introduce non-sequential constraints into the pose graph. In radar SLAM, this task is challenging because radar scans often contain speckle noise, multipath reflections, repeated structures, and viewpoint-dependent intensity changes. Therefore, relying on a single similarity measure can lead to false loop closures. To improve robustness, the proposed system uses a multi-stage loop closure detection and verification pipeline. Candidate loop closures are first retrieved using a compact radar descriptor, then filtered using temporal separation, frequency-domain phase correlation, scan similarity, and geometric consistency checks.

For each radar scan, a descriptor is constructed from its frequency-domain representation. Given a radar scan image $I_i$, compute its 2D Fourier magnitude spectrum and transform the magnitude image into polar coordinates following:

\begin{equation}
P_i = \operatorname{Polar}(|\mathcal{F}\{I_i\}|)
\end{equation}

where $\mathcal{F}$ denotes the 2D Fourier transform. $P_i$ denote the polar-warped Fourier magnitude representations of the scan $I_i$. The descriptor is based on the Fourier magnitude rather than the original scan image since the magnitude spectrum is less sensitive to translation than the spatial-domain scan. The magnitude spectrum is then transformed into a polar representation, where the radial and angular dimensions represent frequency magnitude variations in polar coordinates. A log-compression step is then applied to reduce the effect of dominant high-energy components:

\begin{equation}
\tilde{P}_i = \log(1 + P_i)
\end{equation}

A compact descriptor is then formed by aggregating statistics over the polar frequency representation. As illustrated in Fig. \ref{fig:descriptors}, the descriptor is constructed by concatenating the mean and standard deviation profiles along the radial axis:

\begin{figure}
    \centering
    \includegraphics[width=3.0in]{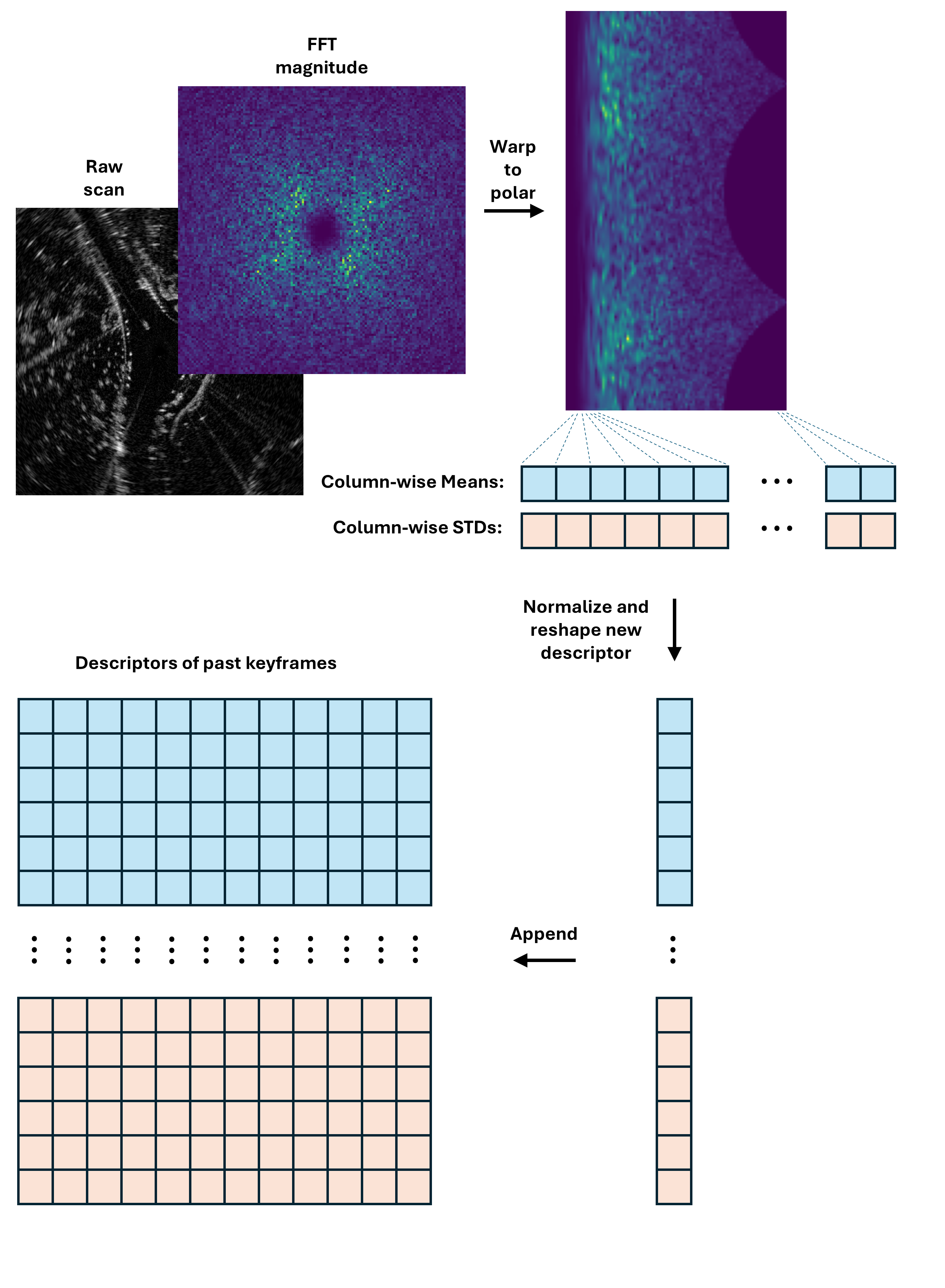}
    \caption{Descriptor generation for loop-candidate retrieval. Each radar scan is transformed into a polar Fourier-magnitude representation, from which column-wise mean and standard deviation profiles are extracted, normalized, concatenated, and stored as compact descriptors for comparison with past keyframes.}
    \label{fig:descriptors}
\end{figure}

\begin{equation}
\mathbf{d}_i =
\begin{bmatrix}
\boldsymbol{\mu}_i \\
\boldsymbol{\sigma}_i
\end{bmatrix}
\end{equation}

where

\begin{equation}
\boldsymbol{\mu}_i = \operatorname{mean}(\tilde{P}_i),
\qquad
\boldsymbol{\sigma}_i = \operatorname{std}(\tilde{P}_i).
\end{equation}

The descriptor is designed to provide a compact representation that is relatively insensitive to scan rotation and translation. A rotation of the original radar scan produces a corresponding circular shift along the angular dimension of the polar frequency representation. Since the descriptor aggregates the polar image using mean and standard deviation profiles over this warped angular dimension, the resulting representation is largely insensitive to such circular shifts. In addition, since the descriptor is computed from the Fourier magnitude spectrum, it is less sensitive to spatial translations of the input scan. These properties provide practical translation and rotation invariance to detect revisits from different angles with reasonable translational offsets (e.g., opposite lanes). In the current implementation, the polar image has size of $360\times91$ pixels, where rows correspond to angle and columns correspond to radial frequency. The column-wise mean and standard deviation profiles therefore produce two vectors of length 91, which are concatenated into a compact descriptor of length 182. The descriptor is finally normalized as:

\begin{equation}
\bar{\mathbf{d}}_i = \frac{\mathbf{d}_i}{\|\mathbf{d}_i\|_2}
\end{equation}

To further reduce the computational cost of loop-closure detection, the system does not invoke a database search for every incoming radar scan. Instead, descriptors are computed and stored for all processed scans, while loop-closure search is triggered only at selected query keyframes. In the current implementation, a scan is used as a loop-closure query only after the vehicle has traveled a minimum distance and the keyframe sampling condition is satisfied. This strategy preserves a dense descriptor database for candidate retrieval while limiting the number of loop-closure verification attempts.

Since all descriptors are $L_2$-normalized, cosine similarity between the current descriptor and any previous descriptor can be computed using a dot product. In practice, however, the descriptor search is performed against all previous keyframes simultaneously. Let $\mathbf{D}_{i-1}$ denote the descriptor database containing the normalized descriptors of all previous scans up to $i-1$:

\begin{equation}   
\mathbf{D}_{i-1}=
    \begin{bmatrix}
    \bar{\mathbf{d}}_0^T \\
    \bar{\mathbf{d}}_1^T \\
    \vdots \\
    \bar{\mathbf{d}}_{i-1}^T
    \end{bmatrix}
\end{equation}

The similarity vector for the current descriptor $\bar{\mathbf{d}}_i$ is then computed using:

\begin{equation}
\mathbf{s}^d_i = \mathbf{D}_{i-1}\bar{\mathbf{d}}_i
\end{equation}

where each element of $\mathbf{s}^d_i$ represents the cosine similarity between the current $i$ and a previous scan $j$:

\begin{equation}
s^d_{ij} = \bar{\mathbf{d}}_j^T \bar{\mathbf{d}}_i
\end{equation}

The top $K$ entries of $\mathbf{s}^d_i$ are selected as initial loop closure candidates:

\begin{equation}
\mathcal{C}_i^0 = \operatorname{TopK}(\mathbf{s}^d_i)
\end{equation}

This matrix-vector formulation allows the current descriptor to be compared with all stored descriptors in a single operation, making the first-stage loop candidate retrieval compact and computationally efficient. For each current scan $i$, the system returns the top $K$ candidates with the highest descriptor similarity, where $K=10$ in the our implementation. This stage is permissive: its purpose is not to make the final loop closure decision, but to generate a small subset of likely candidates for subsequent verification.

After descriptor retrieval, candidates that are too close in time to the current scan are removed. This temporal filtering prevents nearby sequential scans from being incorrectly treated as loop closures. A candidate scan $j$ is rejected if:

\begin{equation}
|i-j| < \Delta_{\min}
\end{equation}

where $\Delta_{\min}$ is the minimum temporal or timestamp-based separation required for a valid loop closure candidate. This ensures that loop closure factors correspond to meaningful revisits rather than local continuity in the trajectory.

The remaining candidates are evaluated using phase-only correlation in the polar frequency domain. Let $P_i$ and $P_j$ denote the polar-warped Fourier magnitude representations of the current and candidate scans $i$ and $j$. The normalized cross-power spectrum is computed and transformed back to the correlation domain as:

\begin{equation}
C_{ij}
=
\mathcal{F}^{-1}
\left\{\frac{\mathcal{F}(P_i)\mathcal{F}(P_j)^{*}}{|\mathcal{F}(P_i)\mathcal{F}(P_j)^{*}|}\right\}
\end{equation}

Where $(\cdot)^{*}$ denotes complex conjugation. The phase-correlation score is then defined as:

\begin{equation}
s_{ij}^{pc} = \max_{\mathbf{v}} C_{ij}(\mathbf{v})
\end{equation}

Where $\mathbf{v}$ indexes locations on the correlation surface. Candidates with phase correlation scores below a threshold $\tau_{pc}$ are rejected according to:

\begin{equation}
s_{ij}^{pc} < \tau_{pc}
\end{equation}

Candidates that pass the phase-correlation test are further evaluated using a scan similarity score after alignment. The aligned candidate scan is compared with the reference scan using a normalized similarity measure. Let $I_i$ denote the reference scan and $I_j^{aligned}$ denote the candidate scan after applying the estimated alignment. The similarity score is computed as:

\begin{equation}
s_{ij}^{a} = \frac{\sum_{\mathbf{u}} I_i(\mathbf{u}) I_j^{aligned}(\mathbf{u})}{\sum_{\mathbf{u}} I_i(\mathbf{u})^2}
\end{equation}

Where $\mathbf{u}$ indexes image pixels. This score measures how well the aligned candidate scan explains the reference scan intensity pattern. Candidates with insufficient alignment similarity are rejected:

\begin{equation}
s_{ij}^{a} < \tau_{a}
\end{equation}

Where $\tau_{a}$ is the alignment similarity threshold. Finally, for each remaining candidate, the relative transform between the two radar scans is estimated using the same procedure described in \cite{NaderFDRio}. The estimated transform is represented as:

\begin{equation}
\mathbf{z}_{ij} =
    \begin{bmatrix}
    \Delta x_{ij} & \Delta y_{ij} & \Delta \theta_{ij}
    \end{bmatrix}^{T}
\end{equation}

This transform defines the relative pose constraint that may be inserted into the pose graph as a loop closure factor. However, before accepting the loop closure, the estimated transform is checked for geometric consistency. A candidate is rejected if the estimated rotation, translation, or final alignment residual is outside physically reasonable limits:

\begin{equation}
|\Delta \theta_{ij}| > \theta_{\max}
\end{equation}

\begin{equation}
\sqrt{\Delta x_{ij}^{2} + \Delta y_{ij}^{2}} > d_{\max}
\end{equation}

or

\begin{equation}
r_{ij} > \tau_{r}
\end{equation}

where $\theta_{\max}$, $d_{\max}$, and $\tau_{r}$ are the maximum allowed rotation, maximum allowed translation, and maximum allowed residuals, respectively. These checks reduce the likelihood of inserting false loop closure factors caused by perceptual aliasing or poor scan alignment.

A loop closure is accepted only if it passes all verification stages. The accepted relative transform $\mathbf{z}_{ij}$ is then added to the pose graph as a non-sequential constraint between poses $i$ and $j$. The complete verification pipeline can be summarized as:

\begin{equation}
\mathcal{C}_i^{0} = \operatorname{TopK}\left(\mathbf{s}_{i}^{d}\right)
\end{equation}

\begin{equation}
\mathcal{C}_i^{1} = \left\{j \in \mathcal{C}_i^{0}\;|\;|i-j| \geq \Delta_{\min}\right\}
\end{equation}

\begin{equation}
\mathcal{C}_i^{2} = \left\{j \in \mathcal{C}_i^{1}\;|\;s_{ij}^{pc} \geq \tau_{pc}\right\}
\end{equation}

\begin{equation}
\mathcal{C}_i^{3} = \left\{j \in \mathcal{C}_i^{2}\;|\;s_{ij}^{a} \geq \tau_{a}\right\}
\end{equation}

\begin{equation}
\begin{aligned}
\mathcal{L}_i =
\Big\{&&j \in \mathcal{C}_i^3 \ \Big| \quad  |\Delta \theta_{ij}|                &&\leq & \quad \theta_{\max}, \\
&& \sqrt{\Delta x_{ij}^{2}+\Delta y_{ij}^{2}} &&\leq & \quad d_{\max}, \\
&& r_{ij}                              &&\leq & \quad \tau_r
\Big\}.
\end{aligned}
\end{equation}

where $\mathcal{C}_i^{0}$ is the initial descriptor-based candidate subset, $\mathcal{C}_i^{1}$ is the temporally filtered subset, $\mathcal{C}_i^{2}$ is the phase-correlation-verified subset, $\mathcal{C}_i^{3}$ is the scan-similarity-verified subset, and $\mathcal{L}_i$ is the final subset of accepted loop closures for scan $i$.

This multi-stage structure is designed to balance recall and precision. The descriptor stage provides efficient retrieval over previously visited scans, while the subsequent verification stages progressively reject candidates that are temporally invalid, poorly correlated in the frequency domain, weakly aligned in the spatial domain, or geometrically inconsistent. As a result, only loop closures with both appearance-level and geometric support are added to the pose graph.

\subsection{Pose Graph Optimization}\label{subsec:pose_graph_optimization}

Graph-based SLAM formulates the trajectory estimation problem as a nonlinear optimization problem over a set of robot poses. In this formulation, each node in the graph represents the vehicle pose at a particular time step, while each edge represents a spatial constraint between two poses. These constraints may come from sequential odometry estimates, loop closure detections, or prior information. The goal of pose graph optimization is to find the configuration of poses that is most consistent with all available constraints. This representation is widely used in SLAM because it separates the sensor-dependent front-end, which constructs the graph, from the sensor-agnostic back-end, which optimizes the graph structure.

In this work, the robot trajectory is represented as a sequence of 2D poses in $SE(2)$. The pose at radar scan $i$ is defined as:

\begin{equation}
    \mathbf{p}_i = \begin{bmatrix} x_i & y_i & \theta_i \end{bmatrix}
    \label{equ:2d_poses}
\end{equation}

where $x_i$ and $y_i$ denote the planar position of the radar sensor and $\theta_i$ denotes the heading angle. The full trajectory is therefore:

\begin{equation}
    \mathcal{X} = \{\mathbf{p}_0, \mathbf{p}_1, \ldots, \mathbf{p}_N\}
    \label{equ:trajectory}
\end{equation}

Each pose can equivalently be represented as a rigid-body transformation:
\begin{equation}
    \mathbf{T}_i =
    \begin{bmatrix}
    \mathbf{R}(\theta_i) & \mathbf{t}_i \\
    \mathbf{0}^{T} & 1
    \end{bmatrix}
    \in SE(2)
    \label{equ:rigid_body_transform}    
\end{equation}

where $\mathbf{R}(\theta_i)$ is the 2D rotation matrix and $\mathbf{t}_i = [x_i, y_i]^T$ is the translation vector.

\begin{figure}[!t]
    \centering
    \includegraphics[width=3.0in]{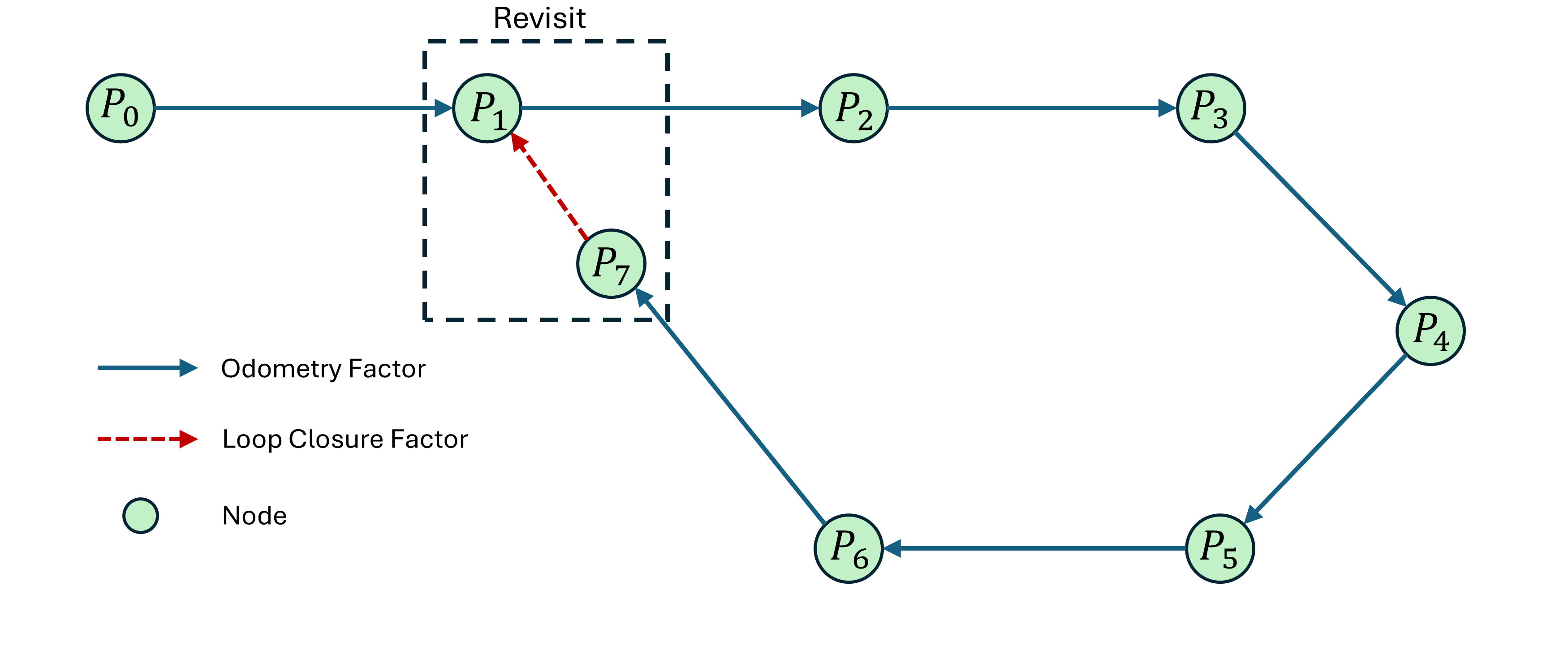}
    \caption{Factor graph representation of FD-SLAM. Sequential odometry factors connect consecutive poses, while verified loop closure factors add non-sequential constraints between revisited locations. The graph is optimized to produce a globally consistent trajectory estimate.}
    \label{fig:pose_graph}
\end{figure}

The graph is constructed from three types of factors. First, a prior factor is added to the initial pose to anchor the graph. Second, odometry factors are added between consecutive poses using the relative motion estimated by the radar odometry front-end. Third, loop closure factors are added between nonconsecutive poses when the radar descriptor search and scan-correlation verification identify a revisited location. In the pose graph, odometry and loop closure constraints are both represented as relative pose measurements between two nodes, although they originate from different sources. A simplified factor graph structure is illustrated in Fig. \ref{fig:pose_graph}, where sequential odometry factors connect consecutive poses and loop closure factors connect non-sequential poses.

For a relative measurement between poses $i$ and $j$, let
\begin{equation}
    \mathit{Z}_{ij} \in SE(2)
    \label{equ:relative_measurement}
\end{equation}

denote the measured relative transform. For odometry factors, $j=i+1$, and $\mathit{Z}_{ij}$ is obtained from consecutive radar scan registration. For loop closure factors, $j$ may be a previously visited nonconsecutive pose, and $\mathit{Z}_{ij}$ is obtained after candidate detection and geometric verification. The predicted relative transform from the current pose estimates is

\begin{equation}
    \hat{\mathit{Z}}_{ij} = \mathbf{T}_i^{-1}\mathbf{T}_j
    \label{equ:predicted_relative_transform}
\end{equation}

The residual error associated with this constraint is computed on the Lie algebra of $SE(2)$:

\begin{equation}
    \mathbf{e}_{ij}
    =
    \operatorname{Log}
    \left(
    \mathit{Z}_{ij}^{-1}
    \hat{\mathit{Z}}_{ij}
    \right)
    \label{equ:residual}
\end{equation}

where $\operatorname{Log}{(\cdot)}$ maps the transformation error from $SE(2)$ to a minimal vector representation in $\mathbb{R}^{3}$. This residual expresses the disagreement between the measured relative motion and the relative motion implied by the current pose estimates.

The weighted nonlinear least-squares estimate of the trajectory is obtained by minimizing the weighted sum of squared residuals:

\begin{equation}
    \begin{aligned}
    \mathcal{X}^{*}
    =
    \arg\min_{\mathcal{X}}
    \Bigg[
    &\sum_{(i,j)\in \mathcal{E}_{loop}}
    \|\mathbf{e}_{ij}^{loop}\|_{\mathbf{\Omega}_{loop}}^{2} \\
    &+
    \sum_{(i,j)\in \mathcal{E}_{odom}}
    \|\mathbf{e}_{ij}^{odom}\|_{\mathbf{\Omega}_{odom}}^{2} \\
    &+
    \|\mathbf{e}_{0}\|_{\mathbf{\Omega}_0}^{2}
    \Bigg]
    \end{aligned}
\end{equation}

where $\mathcal{E}_{odom}$ is the set of odometry edges, $\mathcal{E}_{loop}$ is the set of loop closure edges, and $\mathbf{e}_{0}$ is the residual associated with the prior factor on the initial pose. The notation:

\begin{equation}
    \|\mathbf{e}\|_{\mathbf{\Omega}}^{2}
    =
    \mathbf{e}^{T}\mathbf{\Omega}\mathbf{e}
    \label{equ:residual_norm}
\end{equation}

denotes the Mahalanobis norm, where $\mathbf{\Omega}$ is the information matrix, usually defined as the inverse of the measurement covariance matrix. This weighting allows different confidence levels to be assigned to the prior, odometry constraints, and loop closure constraints.

The pose graph optimizer then redistributes the accumulated odometry error across the trajectory, producing globally more consistent pose estimates. The optimized poses are subsequently used for trajectory evaluation and radar map generation. In our implementation, the pose graph is built and optimized using GTSAM\cite{gtsam}.

\section{Evaluation}\label{sec:evaluation}
\subsection{Qualitative Analysis}\label{subsec:qualitative_analysis}

\begin{figure*}[!t]
    \centering
    \subfloat[]{\includegraphics[height=1.8in, width=2.125in]{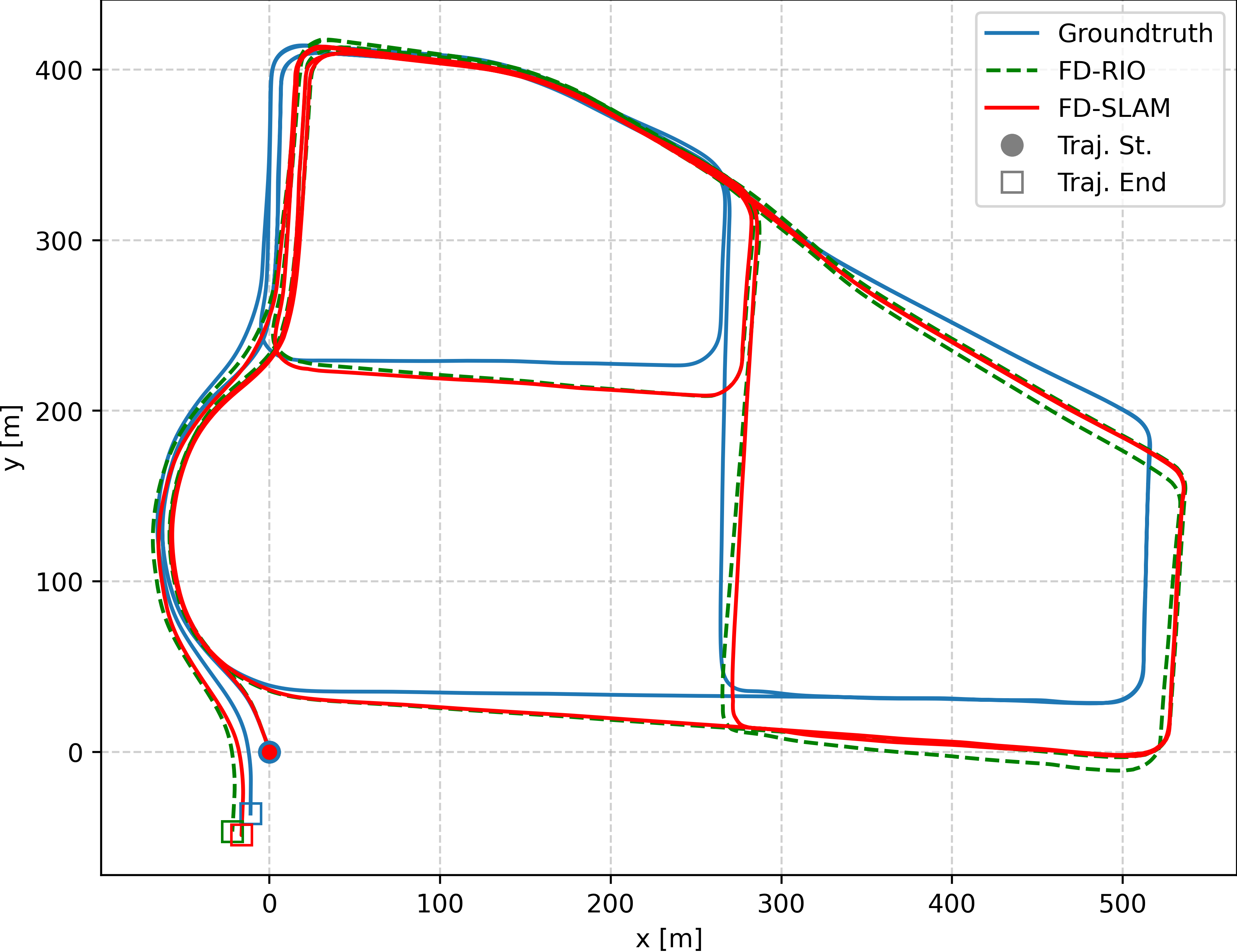}\label{fig:mulran_f1}}\hfil          
    \subfloat[]{\includegraphics[height=1.8in, width=2.125in]{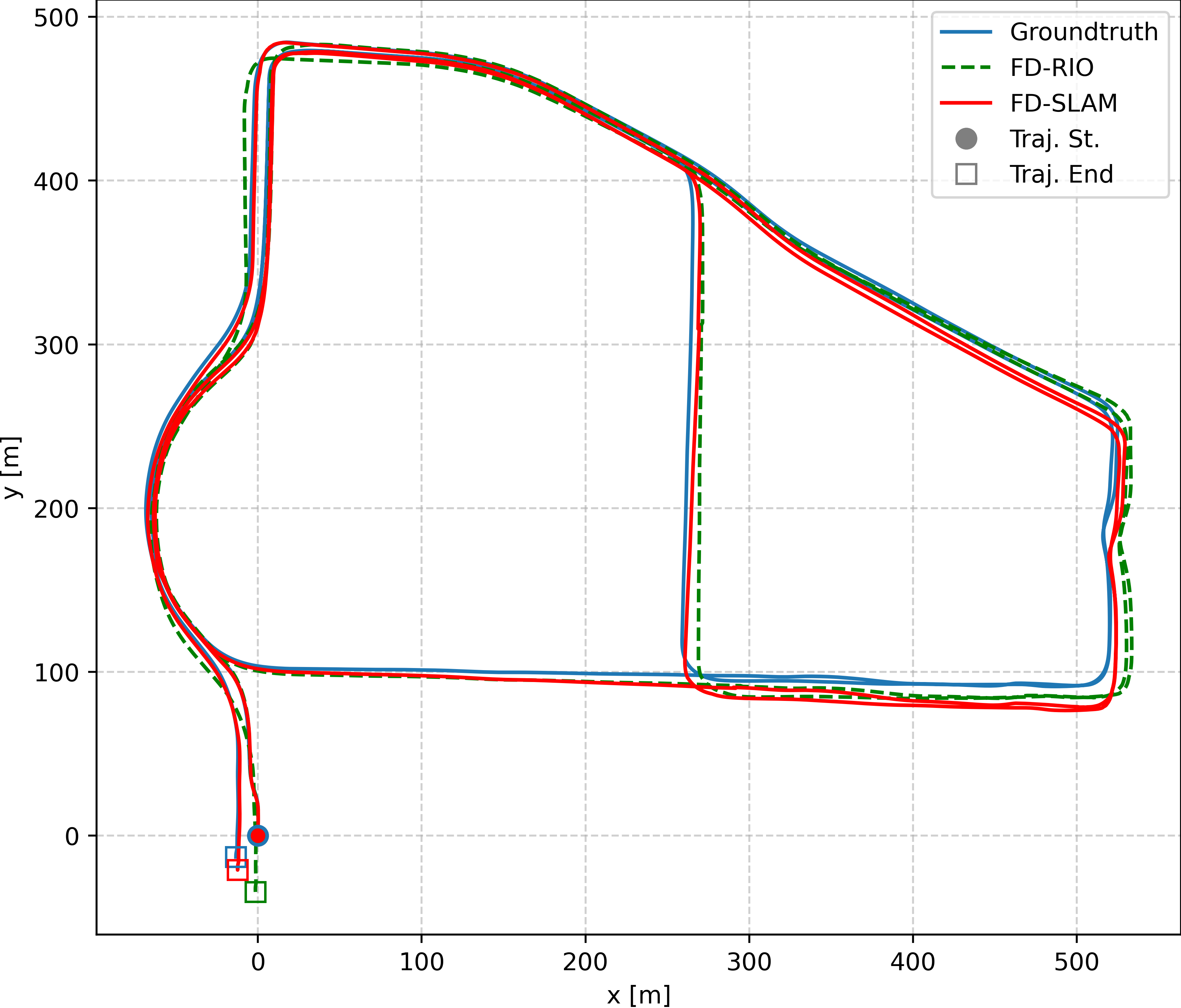}\label{fig:mulran_f2}}\hfil          
    \subfloat[]{\includegraphics[height=1.8in, width=2.125in]{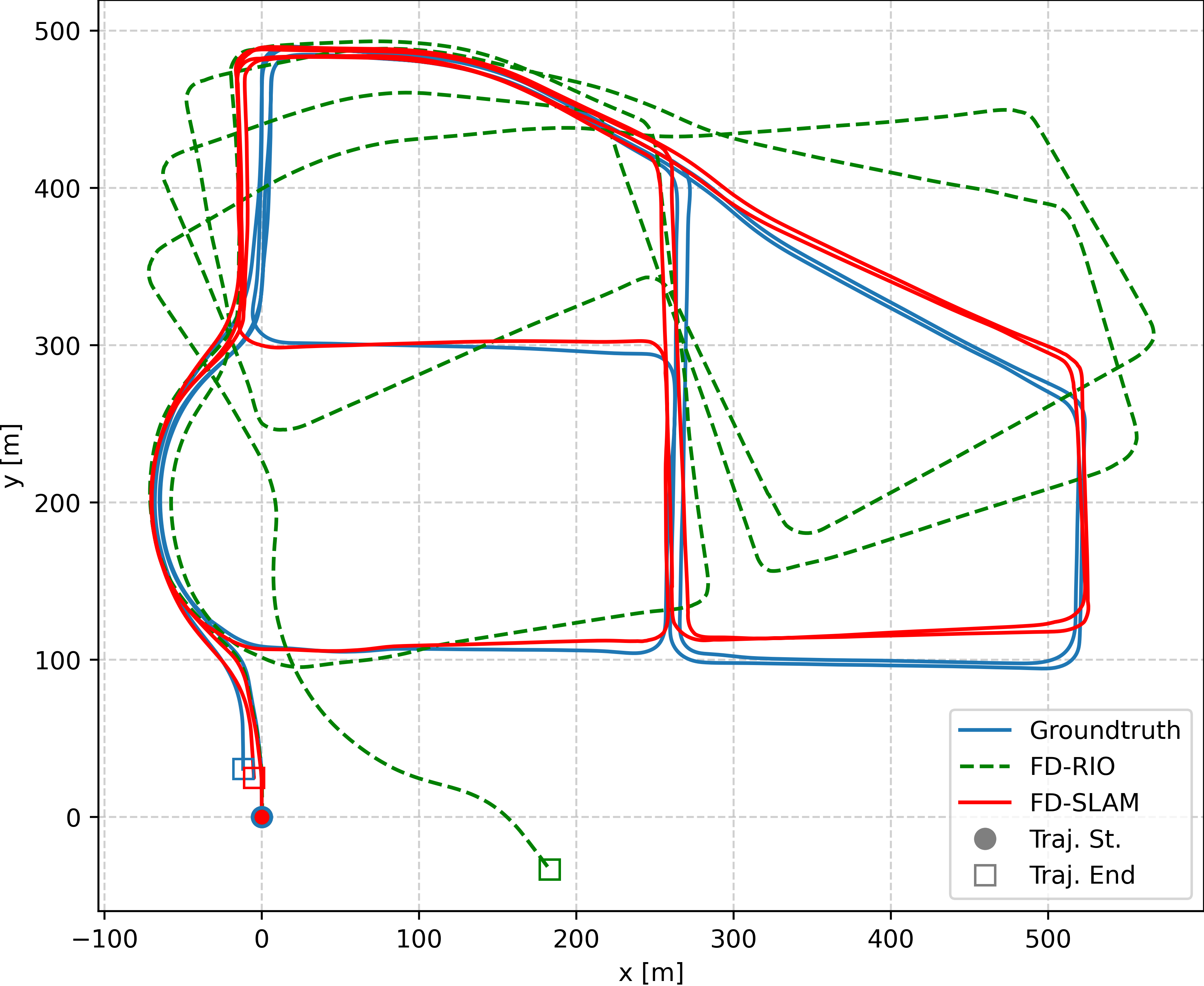}\label{fig:mulran_f3}}\hfil          
    \subfloat[]{\includegraphics[height=1.8in, width=2.125in]{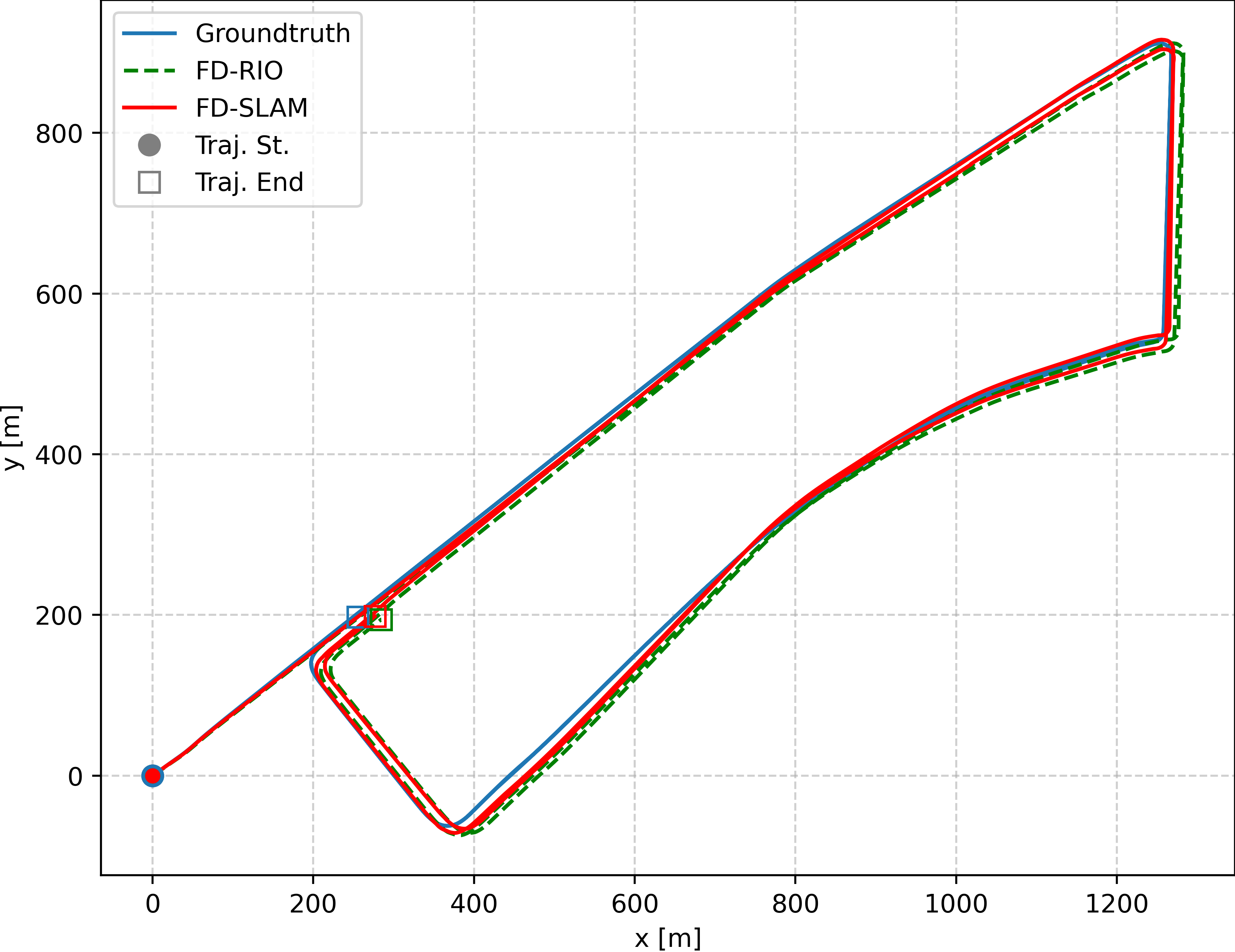}\label{fig:mulran_f4}}\hfil    
    \subfloat[]{\includegraphics[height=1.8in, width=2.125in]{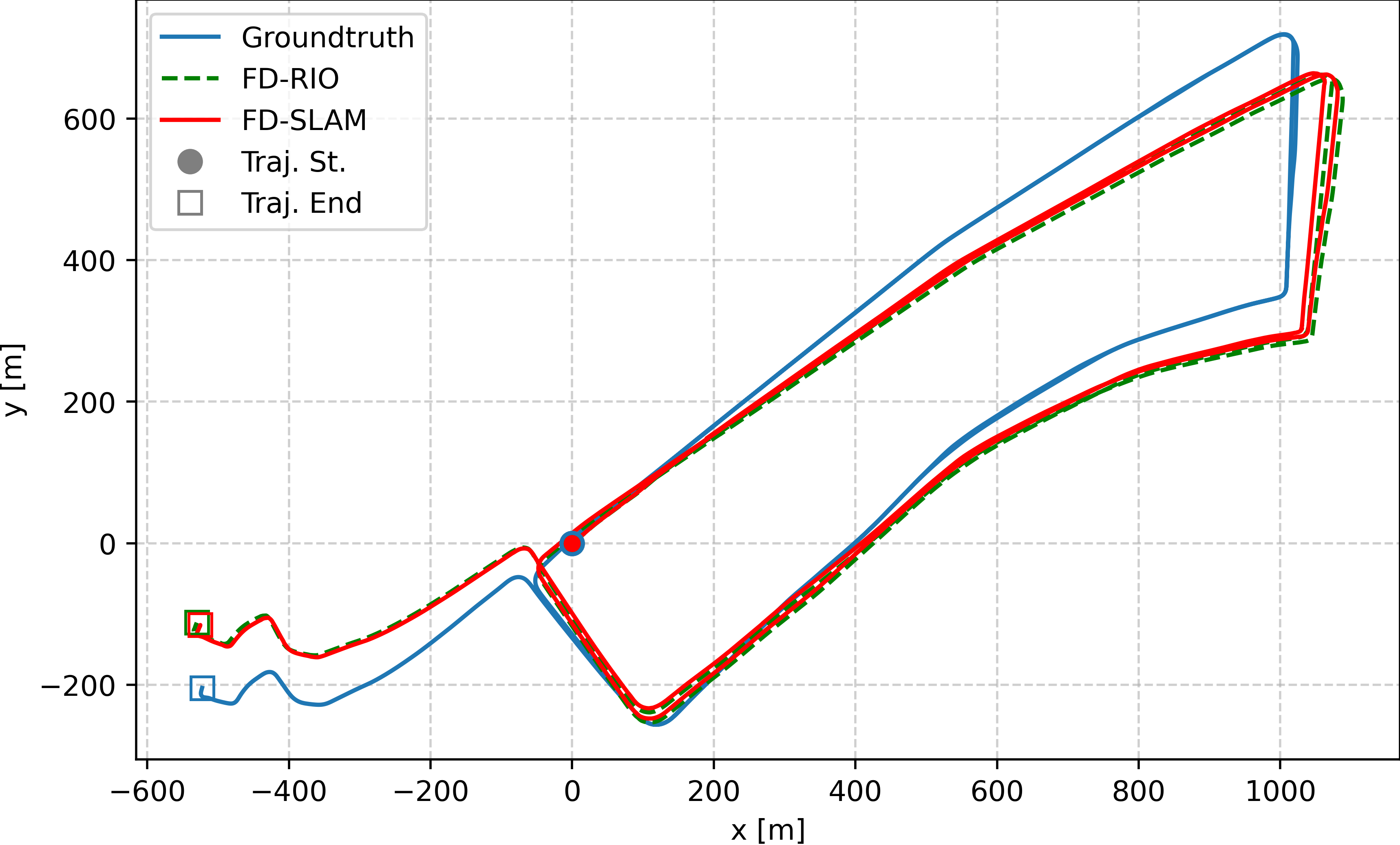}\label{fig:mulran_f5}}\hfil    
    \subfloat[]{\includegraphics[height=1.8in, width=2.125in]{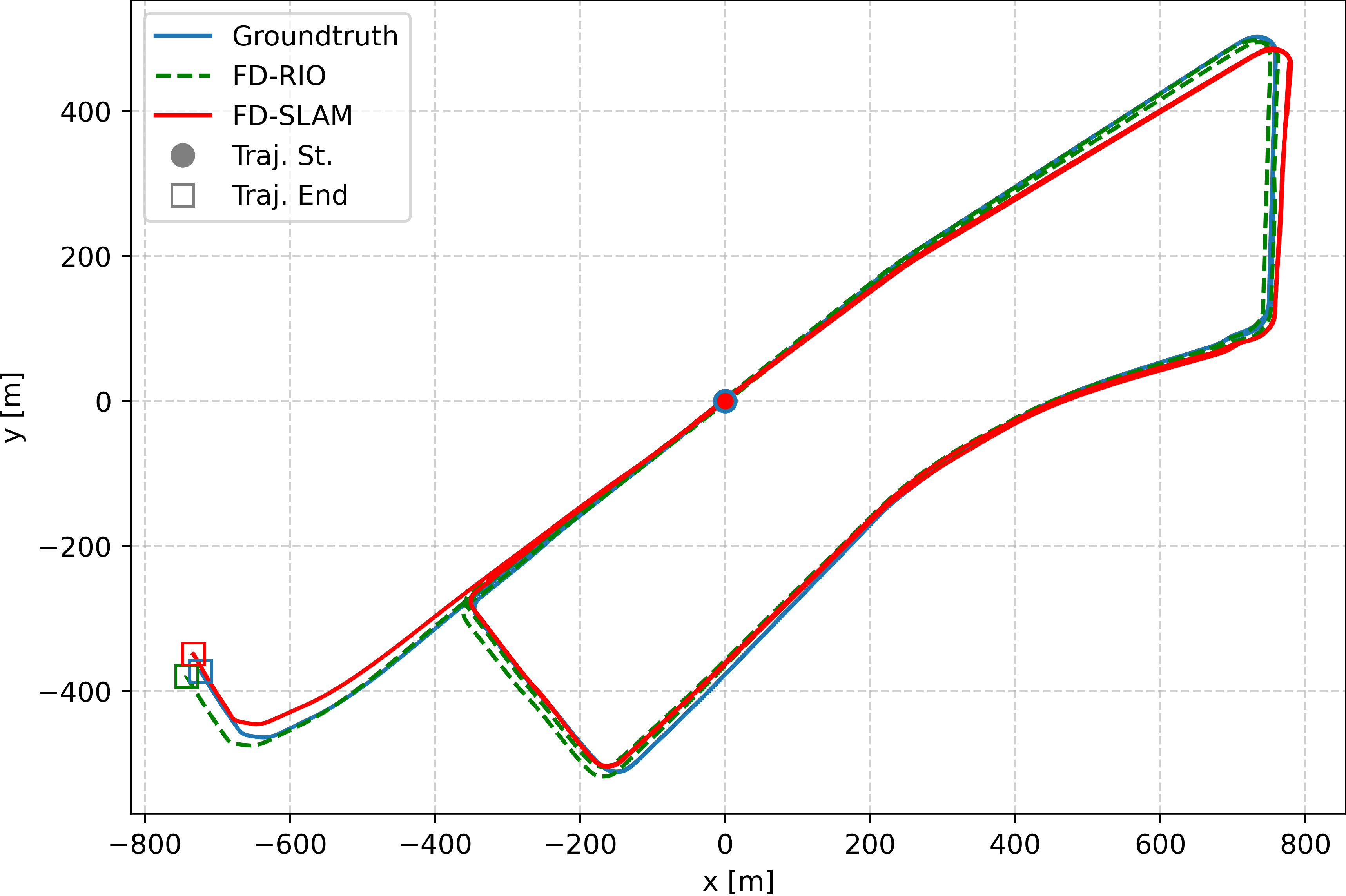}\label{fig:mulran_f6}}\hfil    
    \subfloat[]{\includegraphics[height=1.9in, width=2.4in]{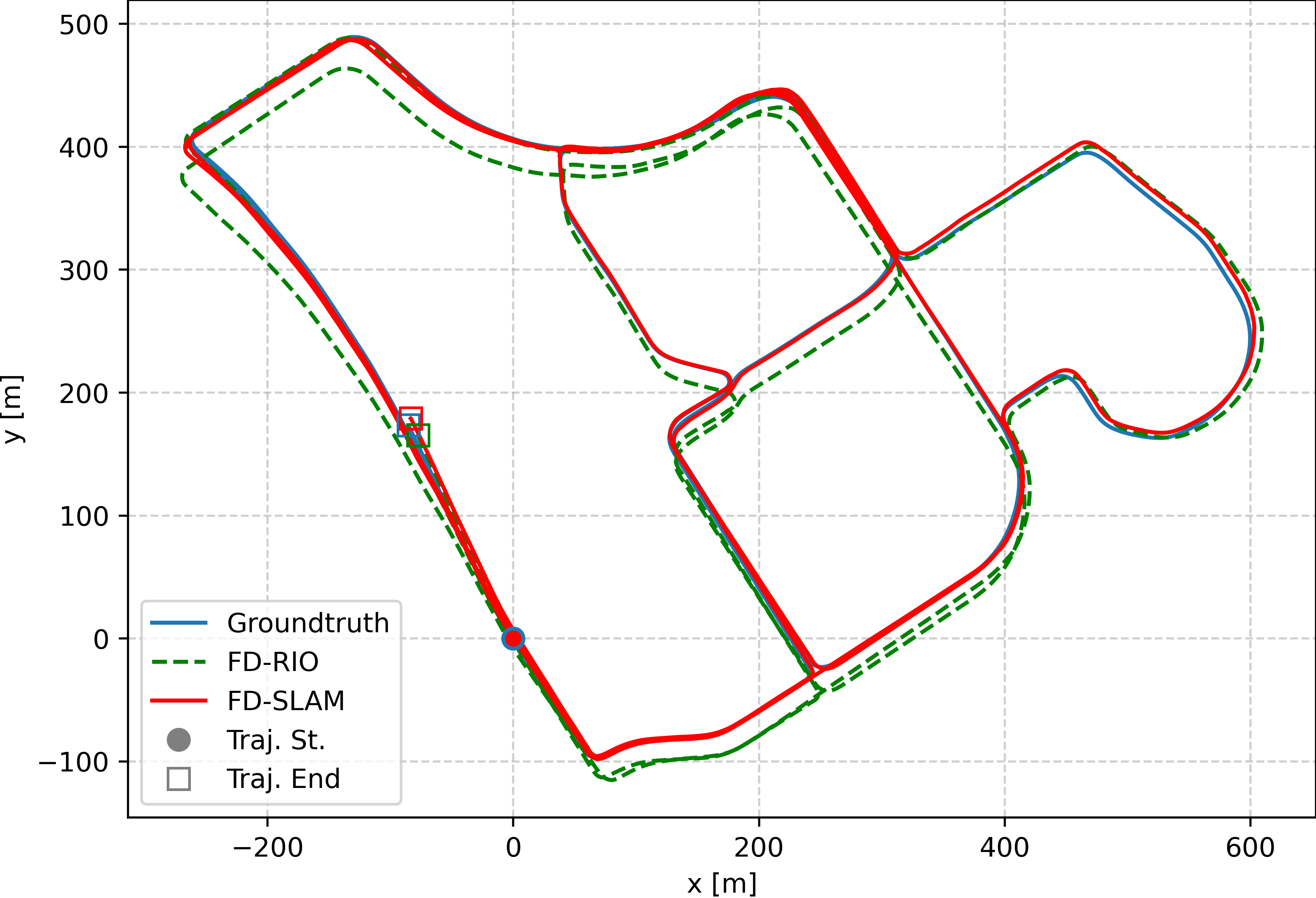}\label{fig:mulran_f7}}\hfil        
    \subfloat[]{\includegraphics[height=1.9in, width=2.4in]{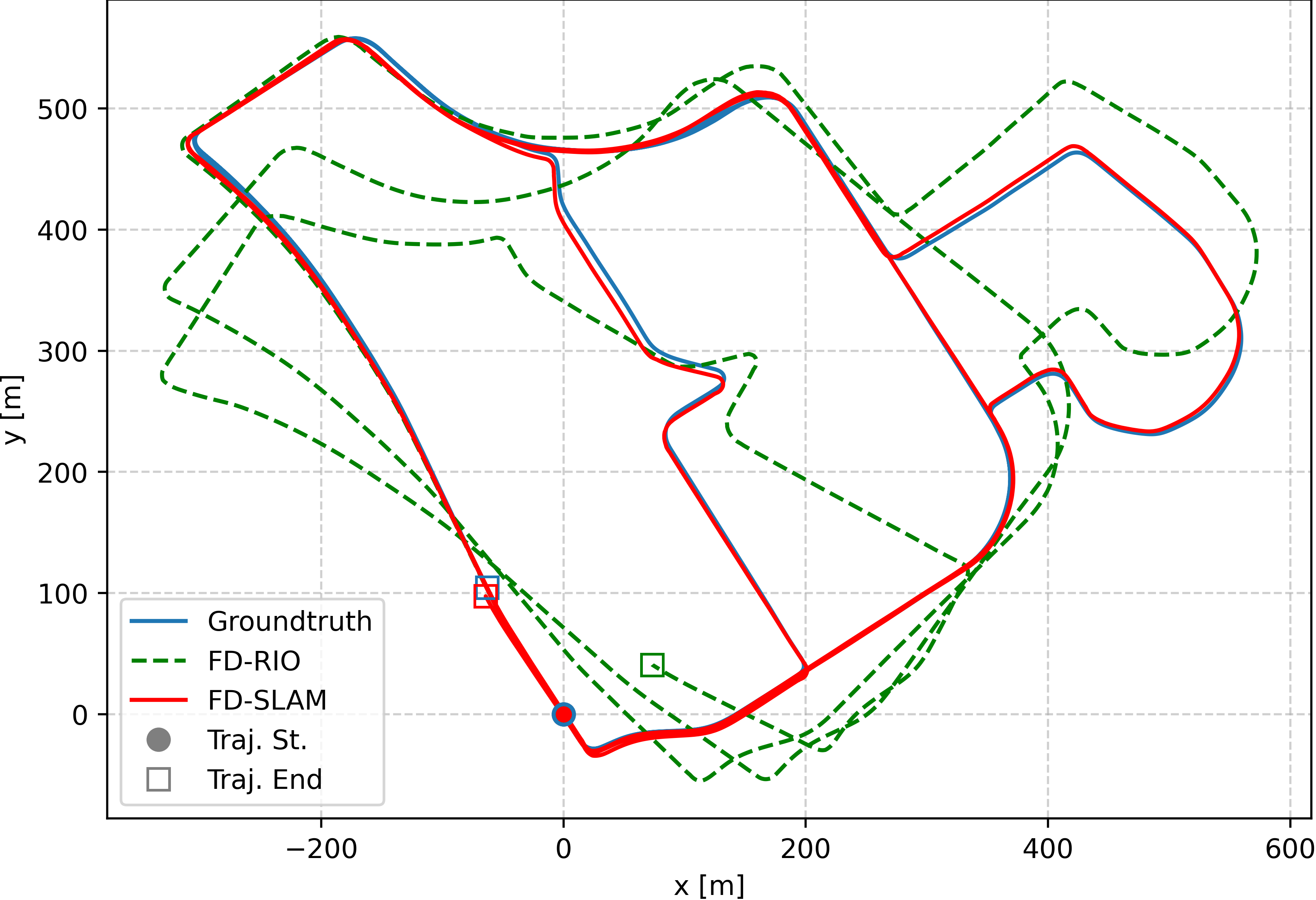}\label{fig:mulran_f8}}\hfil        
    \subfloat[]{\includegraphics[height=1.9in, width=2.4in]{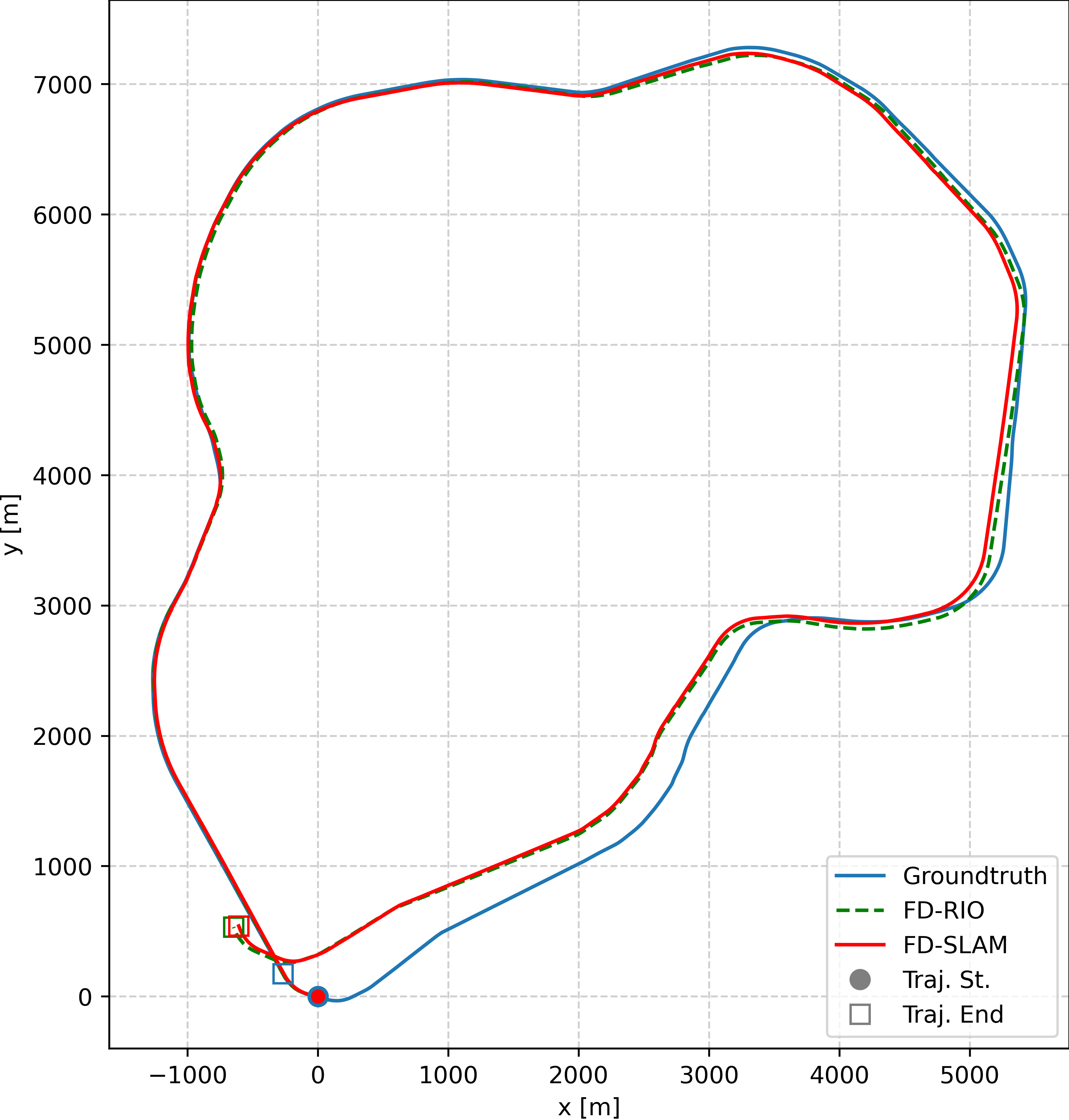}\label{fig:mulran_f9}}\hfil       
    \subfloat[]{\includegraphics[height=1.9in, width=2.4in]{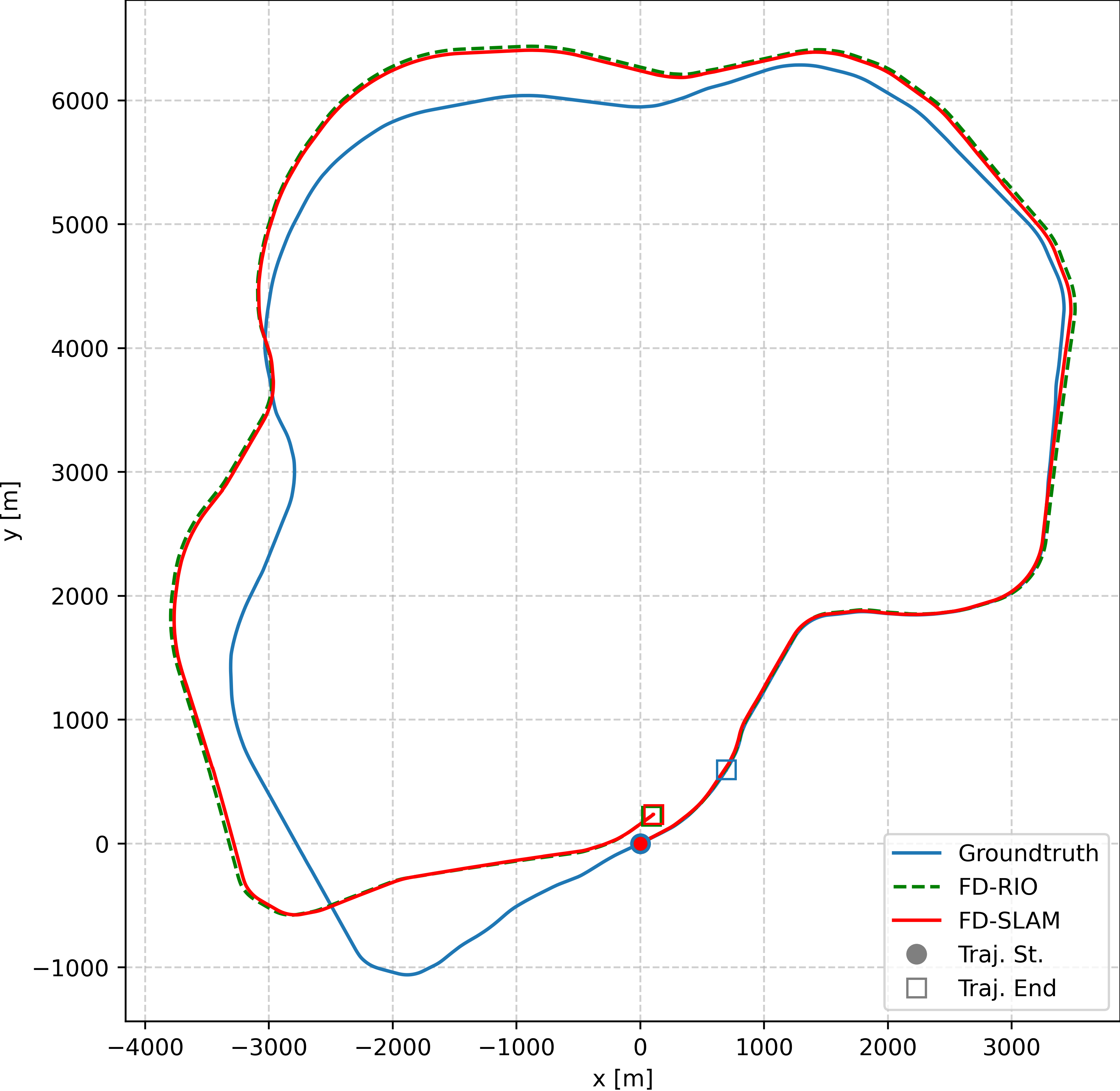}\label{fig:mulran_f10}}\hfil      
    \caption{Qualitative trajectory comparison on ten \textit{MulRan} sequences. Each subplot shows the ground truth, the odometry-only FD-RIO trajectory, and the optimized FD-SLAM trajectory. The optimized trajectories generally preserve the local structure of the radar-inertial odometry estimate while improving global consistency through verified loop closure constraints and pose graph optimization. Sequences shown are: (a) \textit{DCC01}. (b) \textit{DCC02}. (c) \textit{DCC03}. (d) \textit{Riverside01}. (e) \textit{Riverside02}. (f) \textit{Riverside03}. (g) \textit{KAIST02}. (h) \textit{KAIST03}. (i) \textit{Sejong02}. (j) \textit{Sejong03}} 
    \label{fig:mulran_trajectories}
\end{figure*}

Figure \ref{fig:mulran_trajectories} provides a qualitative comparison between the ground-truth trajectories, the odometry-only estimates produced by FD-RIO, and the optimized trajectories produced by the proposed FD-SLAM system on ten \textit{MulRan} sequences. The odometry-only trajectories show the accumulated output of the radar-inertial front-end before loop closure correction, while the optimized trajectories are obtained after inserting verified loop closure constraints and optimizing the resulting pose graph.

The visual results show that the proposed pose graph optimization improves the global consistency of the estimated trajectories while preserving the local trajectory shape produced by the odometry front-end. This behavior is expected because the sequential odometry factors constrain local motion between consecutive poses, whereas loop closure factors introduce long-range constraints that redistribute accumulated drift across the graph. As a result, the optimized trajectory does not simply replace the odometry estimate; instead, it corrects global drift while maintaining the locally consistent structure of the radar-inertial odometry solution.

The benefit of loop closure correction is most visible in sequences where the odometry-only trajectory exhibits substantial drift or distortion. For example, in sequence \textit{DCC03} shown in Fig. \ref{fig:mulran_f3}, the odometry-only result deviates significantly from the ground-truth route and produces a visibly inconsistent trajectory. After optimization, the estimated trajectory is corrected toward the global route structure and becomes substantially more consistent with the ground truth. A similar trend is observed in \textit{KAIST03}, Fig. \ref{fig:mulran_f8}, where the odometry-only estimate shows noticeable drift and trajectory deformation, while the optimized FD-SLAM result better follows the overall shape of the ground-truth trajectory.

In sequences where FD-RIO already provides a locally and globally accurate estimate, the qualitative difference after optimization is smaller. This is visible in Fig. \ref{fig:mulran_f2}, Fig. \ref{fig:mulran_f4}, Fig. \ref{fig:mulran_f6}, and Fig. \ref{fig:mulran_f7}, where the odometry-only and optimized trajectories are already close over large portions of the route. In these cases, the pose graph optimization acts mainly as a refinement step, slightly adjusting the trajectory to improve consistency with revisited locations rather than producing a large correction. This is a desirable behavior, since unnecessary deformation of an already accurate odometry estimate could degrade local consistency.

Some residual errors remain after optimization, especially in sequences with larger drift, limited loop closure availability, or challenging trajectory geometry. For instance, Fig. \ref{fig:mulran_f1}, Fig. \ref{fig:mulran_f5}, and Fig. \ref{fig:mulran_f10} show that while the optimized trajectories are improved relative to the odometry-only estimates, they do not perfectly overlap the ground truth over the entire route. This indicates that the final SLAM accuracy depends not only on the pose graph optimizer but also on the availability, distribution, and quality of accepted loop closure constraints. Incorrect, sparse, or weakly distributed loop closures may limit the ability of the optimizer to fully correct accumulated drift.

\subsection{Quantitative Results}\label{subsec:quantitative_results}

Table I reports the quantitative evaluation of the proposed FD-SLAM system on ten sequences from \textit{MulRan} dataset. The results are reported using the standard KITTI metrics \cite{KITTI}, namely the Average Translational Error and Average Rotational Error, which are calculated by averaging the relative translation error and relative rotation error over sub-trajectories of lengths 100, 200, 300, 400, 500, 600, 700, and 800 m. The resulting values are reported as (translational error [\%] / rotational error [deg/100m]). The open-source toolkit developed by Zhan \textit{et al.} \cite{what_should_be_learnt} was used to run all evaluations. The table includes both radar odometry and radar SLAM methods. Although odometry and SLAM systems do not represent identical problem settings, this comparison provides useful context since FD-SLAM is built on top of the FD-RIO odometry front-end and aims to improve long-term trajectory consistency through loop closure detection and pose graph optimization.

Compared with the FD-RIO odometry baseline, FD-SLAM reduces translational drift on most evaluated sequences. The improvement is particularly clear on \textit{DCC03}, where the translational error decreases from 2.30\% to 1.17\%, and on \textit{KAIST03}, where it decreases from 2.02\% to 1.03\%. FD-SLAM also improves the translational error on \textit{KAIST02}, \textit{Riverside01}, \textit{Riverside02}, \textit{Sejong02}, and \textit{Sejong03}. These results show that the verified loop closure constraints and pose graph optimization are effective in reducing accumulated drift when reliable revisit constraints are available. On some sequences, such as \textit{DCC01} and \textit{DCC02}, the translational change relative to FD-RIO is small, indicating that the benefit of graph optimization depends on the availability, quality, and distribution of accepted loop closures. The highly-repetitive \textit{Riverside} sequences show the smallest overall improvements, with \textit{Riverside03} showing an increase in translational error after optimization where no or bad loop closures were available. 

The rotational results show that FD-SLAM preserves the strong heading performance of the radar-inertial front-end and improves it in several cases. The method achieves the lowest or tied-lowest rotational error on almost all sequences. This behavior is important because heading drift strongly affects long-term trajectory consistency and radar map quality. The results suggest that FD-SLAM help constrain global orientation drift without substantially degrading the local consistency of the FD-RIO front-end.

Among the radar SLAM baselines, TBV SLAM \cite{adolfsson2023tbv} is the strongest competitor and achieves the lowest translational error on several \textit{MulRan} sequences, especially on the \textit{Riverside} routes. However, FD-SLAM provides a fundamentally different design approach here: while TBV is a sparse radar SLAM system based on CFEAR \cite{adolfsson2023cfear} odometry, Scan Context-based retrieval, and learned alignment verification, FD-SLAM preserves the dense scanning-radar representation and uses frequency-domain descriptor retrieval with deterministic correlation-based and geometric verification. Moreover, FD-SLAM does not depend on scans preprocessing, motion distortion compensation, or odometry outlier rejection. Despite these methodological differences, FD-SLAM remains on par in translational accuracy and achieves lower rotational error than TBV on many shared sequences. Finally, compared with RadarSLAM \cite{hong2022radarslam}, FD-SLAM achieves overall stronger performance on all sequences.

Overall, the quantitative results show that FD-SLAM improves the FD-RIO odometry baseline in most sequences and achieves competitive performance against recent radar odometry and radar SLAM methods. The results also demonstrate that dense frequency-domain radar representations can support loop candidate retrieval, loop verification, and pose graph optimization while maintaining competitive trajectory accuracy and favorable rotational drift.

\begin{table*}
    \begin{center}
    
    \caption{Evalution results on ten sequences from \textit{MulRan} Dataset}\label{tab:mulran_results}
    \vspace{-8pt}
    \resizebox{\linewidth}{!}{%
    \begin{tabular}{c|l|c|cccccccccc}
        \hline
           &         &     & \multicolumn{10}{c}{\textbf{Sequence}}            \\
        \textbf{Method} & \textbf{Method} &  \textbf{Type} & \textbf{\textit{DCC01}} &  \textbf{\textit{DCC02}} &  \textbf{\textit{DCC03}} & \textbf{\textit{KAIST02}} & \textbf{\textit{KAIST03}} & \textbf{\textit{Riverside01}} & \textbf{\textit{Riverside02}} & \textbf{\textit{Riverside03}} & \textbf{\textit{Sejong02}} & \textbf{\textit{Sejong03}}  \\ \hline \hline

      \multirow{4}{*}{\adjustbox{angle=45}{\textbf{Odometry}}} & CFEAR-3-s50 \cite{adolfsson2023cfear}  & S &   2.09/0.55   &   1.38/0.47  &1.26/0.47  &   1.51/0.63   &   1.59/0.75   &   1.62/0.62   &   1.35/0.52  &   1.19/0.37   &   -           &   -            \\
       & SDRO \cite{zhang2023sdr}                             & S &\e{1.55}/0.35  &   1.53/0.33   & 1.60/0.30 &   1.61/0.35   & 1.59/0.32 & 1.61/\e{0.26} & 1.59/\e{0.27} &   1.62/0.29   &   -           &   -            \\
       & $R^3O$ \cite{r3o}                                    & D &   2.39/0.43   &   1.40/0.34   &   1.48/0.41   &   1.55/0.53   &1.53/0.50  & 1.34/0.39  &   1.98/0.53   &   1.81/0.57   &   -           &   -            \\
       & FD-RIO \cite{NaderFDRio}                             & D & 2.13/\e{0.21} & 1.49/\e{0.23} &   2.30/0.69   &1.16/\e{0.24} &   2.02/0.54&   1.50/0.38   &   1.46/0.31   &1.08/\e{0.13}&1.44/\e{0.26}&2.21/\e{0.48} \\ \hline
      \multirow{3}{*}{\adjustbox{angle=45}{\textbf{SLAM}}} & SuMa \cite{behley2018suma,hong2022radarslam} & (Lidar) &   2.71/0.40  &   4.07/0.90   &   2.14/0.60   &   2.64/0.60   &   2.17/0.60   &     -       &     -    &    -   &   -   &   -  \\
       & RadarSLAM \cite{hong2022radarslam}                   & S &   2.39/0.40   &   1.90/0.40   &   1.56/0.20   &   1.76/0.40   &   1.72/0.40   &   3.40/0.90   &   1.79/0.30   &   1.95/0.50   &   -           &   -            \\
       & TBV SLAM \cite{adolfsson2023tbv}                     & S &   2.01/0.27   &   \e{1.35}/0.25   &   \e{1.14}/0.22   &   \e{1.03}/0.30   &   1.08/0.35   &   \e{1.25}/0.35   &   \e{1.09}/0.30   &   \e{0.99}/0.18   &   -           &   -            \\
       & FD-SLAM (Ours)                                       & D & 2.10/\e{0.21}     & 1.39/\e{0.23}     &   1.17/\e{0.18}  &   1.11/\e{0.24}   &   \e{1.03}/\e{0.25}   &   1.43/0.37    &   1.39/0.31   &   1.39/0.14  &   \e{1.31}/\e{0.26}   &   \e{2.02}/\e{0.48}    \\ \hline

            \multicolumn{13}{l}{Standard KITTI evaluation metrics are used; translational error [\%] / rotational error [deg/100m]. (-) indicates results that are not available or not applicable.}\\
            \multicolumn{13}{l}{(S) stands for Sparse method, (D) is for Dense method.}\\
            \end{tabular}%
            }
\end{center}
\end{table*}

\subsection{Scan Matching and Descriptor Evaluation}\label{subsec:descriptor_evaluation}

To evaluate the effectiveness of the proposed radar descriptor for loop closure search, we analyzed the relationship between descriptor similarity and the true spatial separation between scan pairs. The goal of this experiment is to verify that the descriptor + cosine similarity are suitable for candidate retrieval, i.e., for narrowing the search space before applying the remaining verification stages described earlier.

Three sequences were used in this test: \textit{DCC01}, \textit{KAIST03}, and \textit{Riverside01}. Using the corresponding ground truth poses, the Euclidean distance between all scan pairs was calculated to quantify their true spatial separation. In parallel, the cosine similarities between the descriptors of all scan pairs were computed using the descriptor database matrix $\mathbf{D}$ defined earlier. The full pairwise similarity matrix for each sequence can be obtained using:
\begin{equation}
    \mathbf{\mathcal{S}} = \mathbf{D}\mathbf{D}^T
\end{equation}

Each element of $\mathbf{\mathcal{S}}$ gives the cosine similarity between a pair of radar scans: $\mathcal{S}_{ij} = \bar{\mathbf{d}}_i^T \bar{\mathbf{d}}_j$. This similarity matrix is then compared with the ground-truth pairwise spatial distances to evaluate whether descriptor similarity decreases as physical separation increases.

\begin{figure*}[!t]
    \centering
    \includegraphics[width=7.0in]{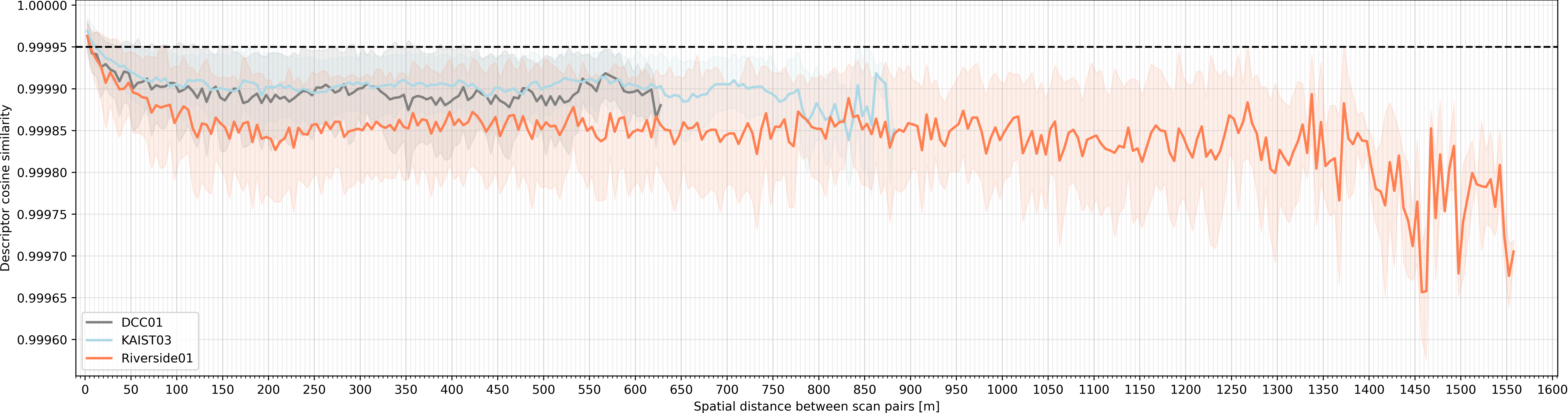}
    \caption{Descriptor similarity as a function of spatial separation for scan pairs from three \textit{MulRan} sequences. The spatial distance between each pair of scans is computed from ground-truth poses, while descriptor similarity is computed using cosine similarity. The plotted curves show the binned mean similarity values, and the shaded regions indicate one standard deviation within each distance bin. The three sequences have different trajectory lengths, which results in different maximum spatial distances. The dashed horizontal line illustrates a candidate-retrieval threshold: only scan pairs with similarity above this threshold are retained as high-likelihood candidates, allowing the loop-closure search to be narrowed to the closest top-$K$ descriptor matches.}
    \label{fig:descriptor_validation}
\end{figure*}

Figure \ref{fig:descriptor_validation} shows the descriptor cosine similarity as a function of the ground-truth spatial distance between scan pairs for the three evaluated sequences. In all cases, scan pairs that are spatially closer tend to exhibit higher descriptor similarity, while the similarity generally decreases as the spatial separation increases. Although the absolute variation in similarity is small, the trend is consistent across the tested sequences and indicates that the descriptor retains meaningful place information with sufficient translational and rotational invariance. This behavior is desirable for loop-closure search, since it suggests that scans acquired from nearby or revisited locations are more likely to appear among the most similar candidates. The figure shows that the descriptor achieves its goal of keeping loop-closure search computationally tractable; even though descriptor similarity alone is not sufficient to guarantee a correct loop closure, it provides a reliable way to prioritize plausible matches and narrow the search to a small subset of high-probability candidates. This is especially important for long trajectories, where exhaustive pairwise scan matching would be computationally expensive. Therefore, the descriptor contributes to the overall loop-closure pipeline primarily by improving search efficiency while preserving enough discriminative power to retrieve relevant candidate matches.

\subsection{Runtime Analysis}\label{subsec:runtime}

The computational performance of the proposed system was evaluated to assess its suitability for real-time radar SLAM. All experiments were performed on an office PC equipped with an Intel(R) i7-13700 CPU at 2.10 GHz and 32 GB of RAM, without GPU acceleration. The timing analysis separates the foreground radar-inertial odometry front-end from the background SLAM processes. This separation is important because the front-end must process incoming radar and IMU measurements at sensor rate, while loop closure search and pose graph optimization can be executed asynchronously.

\begin{figure}[!t]
    \centering
    \includegraphics[width=3.0in]{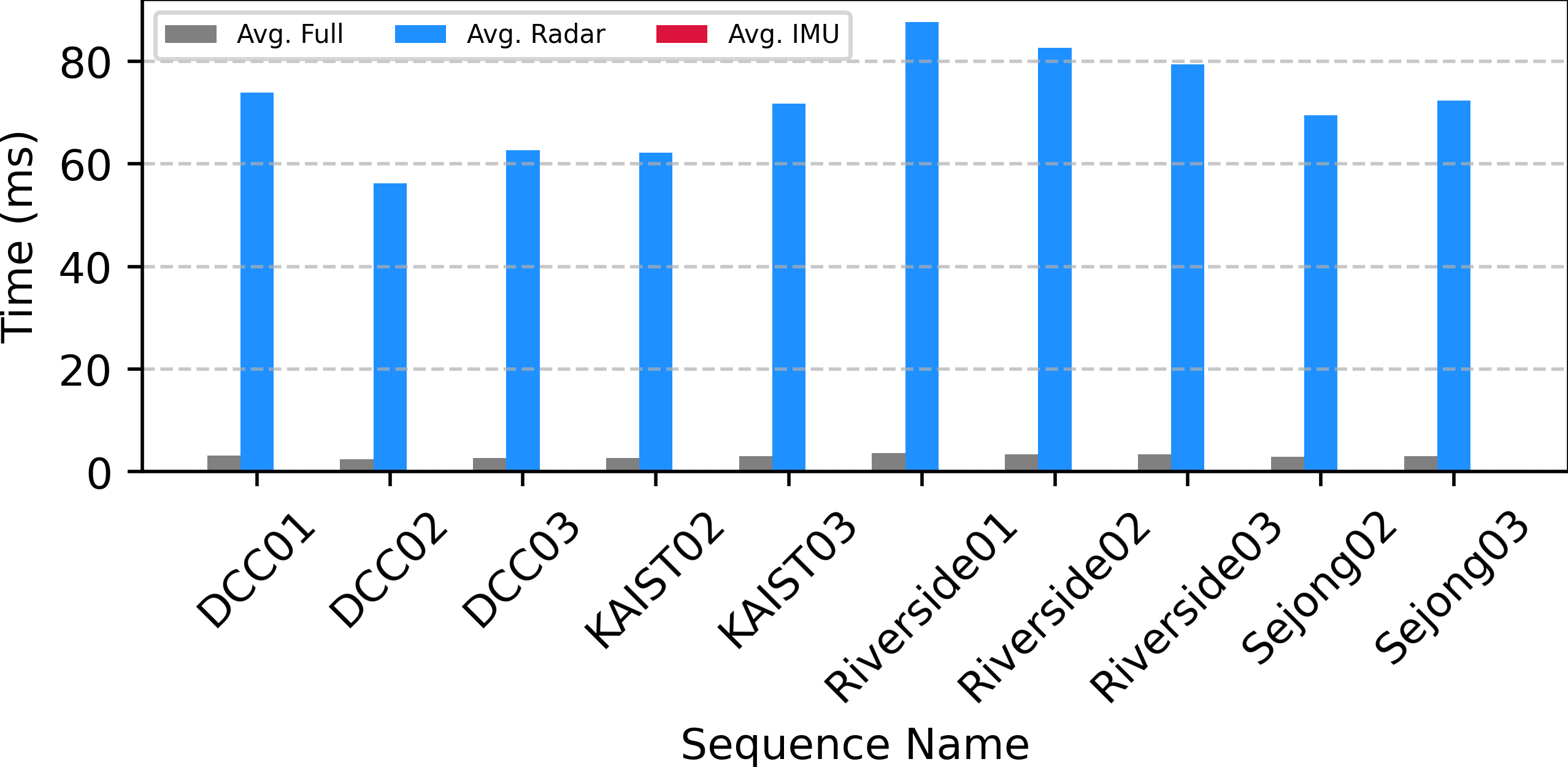}
    \caption{Average foreground processing time across the evaluated \textit{MulRan} sequences. Radar scan processing dominates the front-end runtime, requiring approximately $70\,\mathrm{ms}$ per scan on average which corresponds to an effective radar processing rate of approximately $14\,\mathrm{Hz}$. IMU processing is negligible on this scale.}
    \label{fig:avg_times}
\end{figure}

Figure \ref{fig:avg_times} shows the average processing time of the foreground odometry front-end for each \textit{MulRan} sequence. Radar scan processing is the dominant foreground cost, with an average processing time of approximately $70~\mathrm{ms}$ per scan across the tested sequences. This corresponds to an effective radar processing rate of approximately $14~\mathrm{Hz}$. Since the scanning radar in the \textit{MulRan} dataset operates at approximately $4~\mathrm{Hz}$, the front-end processes radar scans substantially faster than the sensor sampling rate. IMU updates require negligible computation time compared with radar scan processing, remaining close to zero on the millisecond scale shown in the figure. The average full front-end update time remains low because only a fraction of updates correspond to radar frames, while the high-rate IMU updates are computationally lightweight.

\begin{figure}[!t]
    \centering
    \includegraphics[width=3.0in]{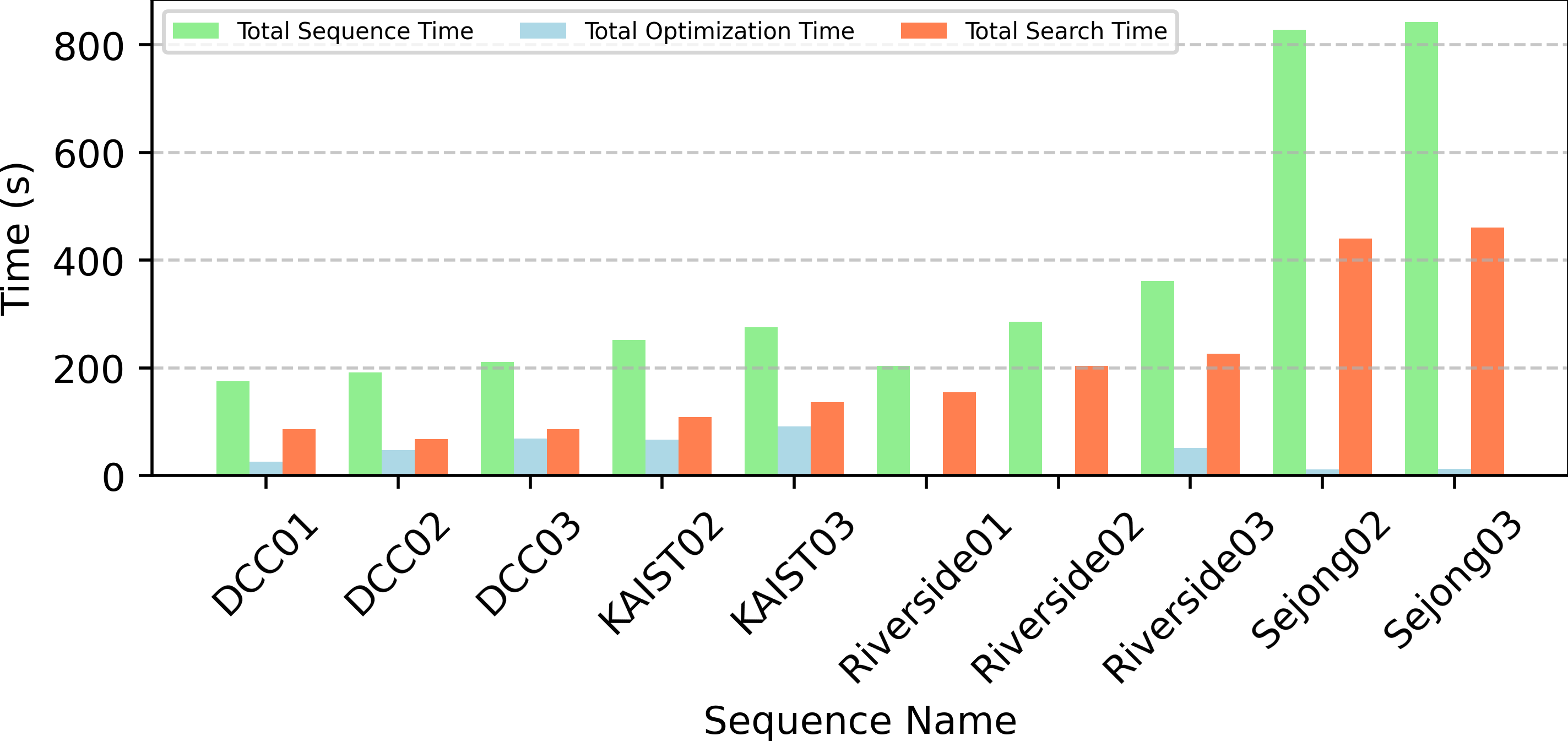}
    \caption{Runtime breakdown of the background SLAM components across the evaluated \textit{MulRan} sequences. The total loop closure search time and pose graph optimization time are shown next to the total sequence processing time. Search is the dominant background cost, while optimization remains comparatively small. The results support parallel execution of loop closure detection and graph optimization alongside the real-time odometry front-end.}
    \label{fig:seq_optim_tot}
\end{figure}

Figure \ref{fig:seq_optim_tot} reports the accumulated runtime of the background SLAM components for each evaluated sequence. The figure compares the total time required to process a full sequence with the total time spent in loop closure search and pose graph optimization. The loop closure search is the dominant background cost, while pose graph optimization requires substantially less time in most sequences. This is expected because the search stage evaluates candidate revisits over the stored descriptor database and performs verification steps, whereas optimization is triggered only when loop closure constraints are added to the graph.

The background timing results support the intended asynchronous design of the system. In all tested sequences, the combined time required for loop closure search and graph optimization remains below the total sequence processing time. This indicates that these operations can be executed in parallel with the foreground odometry front-end without necessarily blocking sensor processing. The front-end maintains real-time local odometry, while global correction is performed asynchronously whenever loop closure candidates are detected and verified. Therefore, the relevant computational requirement is that the background processes do not accumulate an unbounded backlog over the sequence. The results in Fig. \ref{fig:seq_optim_tot} suggest that, for the evaluated \textit{MulRan} sequences, the background pipeline is computationally feasible for asynchronous execution on the tested CPU-only setup. The longer \textit{Sejong} sequences require more total search time, as expected from their larger trajectory length and larger descriptor database. Nevertheless, the search and optimization costs remain compatible with a background processing model.

Overall, the runtime results show that the radar-inertial odometry front-end operates comfortably above the radar sampling rate, while the loop closure and optimization components remain suitable for background execution. This supports the use of the proposed FD-SLAM pipeline as a lightweight radar SLAM system capable of real-time local odometry and asynchronous global trajectory correction.

\section{Conclusion}\label{sec:conclusion}

This paper presented FD-SLAM, a dense radar-inertial SLAM system for autonomous ground vehicles using scanning FMCW radar and IMU measurements. The proposed system extends the FD-RIO odometry front-end into a graph-based SLAM pipeline by adding descriptor-based loop closure detection, multi-stage loop verification, and pose graph optimization. A central design choice in FD-SLAM is to preserve the dense radar scan representation throughout the loop closure pipeline. Rather than converting scans into sparse landmarks, point clouds, or visual-feature descriptors, the proposed method uses a compact frequency domain -based descriptor for candidate retrieval and verifies candidates using a multi-staged procedure. This provides a radar-native alternative to sparse feature-based radar SLAM pipelines and demonstrates that dense frequency-domain radar representations can support a complete SLAM system.

The results show that pose graph optimization improves global trajectory consistency while preserving the local structure of the radar-inertial odometry estimate. The quantitative results show that FD-SLAM improves the FD-RIO odometry baseline and achieves competitive performance compared with the current state-of-the-art radar SLAM methods. In particular, our method provides favorable rotational accuracy on several routes, indicating that the verified loop closures and graph optimization help constrain heading drift without degrading local odometry consistency.

Future work will focus on improving loop closure robustness in perceptually ambiguous and repetitive radar scenes, as well as extending the framework toward map reuse and relocalization for long-term autonomous navigation.

\IEEEtriggeratref{27}

\bibliographystyle{IEEEtran}
\bibliography{myref}
\newpage

\end{document}